\documentclass{article}

\usepackage{iclr2027_conference,times}
\iclrfinaltrue
  
\usepackage{amsmath,amsfonts,bm}

\def\eqref#1{equation~\ref{#1}}

\def\1{\bm{1}}

\DeclareMathAlphabet{\mathsfit}{\encodingdefault}{\sfdefault}{m}{sl}
\SetMathAlphabet{\mathsfit}{bold}{\encodingdefault}{\sfdefault}{bx}{n}

\usepackage{hyperref}
\usepackage{url}
\usepackage{booktabs}
\usepackage{graphicx}
\usepackage{subcaption}
\usepackage{multirow}
\usepackage{array}
\usepackage{colortbl}
\usepackage{amsmath}
\usepackage{amssymb}
\usepackage{xspace}
\usepackage{float}
\usepackage{algorithm}
\usepackage{algorithmic}

\title{TIDE: Teacher-Student Transition via Informative Distillation and Exploration for Agentic RL}

\author{
Yibin Huang$^{1}$, Xinming Xu$^{2}$, Conghui Zhu$^{1}$\thanks{Corresponding author.} \\[0.5em]
$^{1}$Faculty of Computing, Harbin Institute of Technology \qquad
$^{2}$Tsinghua University
}

\newcommand{\gap}{D}
\newcommand{\qscore}{\tilde{Q}}
\newcommand{\dscore}{\tilde{D}}
\newcommand{\tide}{\textsc{TIDE}\xspace}

\begin{document}
\maketitle

\fancyhead[L]{}

\begin{abstract}
Effective multi-turn agents require interaction strategies that coordinate information gathering, actions, and feedback over long horizons. GRPO is a reinforcement learning algorithm used to train these agents, but sparse trajectory-level rewards limit early exploration in small models. Recent methods augment RL with on-policy distillation (OPD) from a stronger teacher. However, a fixed mixture assumes that teacher guidance and reward optimization should retain a constant relative role throughout training and across interaction turns. This assumption can fail at two scales. Globally, as training progresses, maintaining strong distillation pressure can constrain the model from moving beyond the teacher's capabilities. Locally, teacher--student disagreement identifies where the student departs from the teacher, but cannot tell whether that departure is exploration supported by better outcomes or low-quality policy drift. Our methodological insight is that teacher guidance and reward optimization should be dynamically rebalanced over training and jointly allocated across turns. We instantiate this insight in \tide. Globally, \tide uses the measured disagreement trend as a practical schedule signal, advancing an OPD-to-RL handoff when discrepancy reduction becomes slow but remains positive and progressively increasing the relative weight of RL. Locally, \tide jointly modulates teacher-guided and reward-driven updates: relative action value and disagreement prioritize the OPD signal, whereas relative action value supplies the RL advantage and normalized disagreement reweights it across turns. Coupled with the global handoff, \tide allocates stronger teacher guidance early and gives reward-driven updates greater relative weight later in training. Experiments across multiple benchmarks, student scales, and controlled ablations support the effectiveness of TIDE's adaptive OPD--RL coordination.
\end{abstract}

\section{Introduction}
\label{sec:introduction}

Language-model agents have emerged as a paradigm for long-horizon tasks such as web navigation and embodied control~\citep{yao2022webshop,shridhar2020alfworld}. In these tasks, actions alter subsequent states but may reveal consequences only after many turns, making effective policy learning from sparse outcome rewards difficult and motivating richer supervision.

GRPO~\citep{shao2024deepseekmath} and on-policy distillation (OPD)~\citep{agarwal2024policy} provide complementary supervision for learning long-horizon interaction policies. GRPO optimizes the policy through group-relative comparisons of trajectory-level task rewards, whereas OPD provides token-level guidance by distilling a stronger teacher's distribution on student-generated trajectories.

Yet combining OPD with RL raises two allocation questions: how should teacher supervision be scheduled across training, and how should learning signals be allocated within each trajectory? At the global scale, a fixed-weight mixture cannot accommodate changing supervision utility. As training progresses, the marginal utility of continued OPD may diminish~\citep{tan2026atod,yu2026retireopd}; retaining a large OPD coefficient in this latter phase can constrain reward-driven departures from the teacher~\citep{yu2026retireopd}.
At the local scale, student rollouts can enter low-quality regions that depart from the teacher distribution, where teacher-provided distillation signals may become less reliable~\citep{zheng2026scope,lin2026policy}. Conversely, some departures from the teacher distribution can be beneficial and yield higher returns~\citep{liu2026teacher,wang2026distilled}.

\begin{figure}[t]
  \centering
  \includegraphics[width=\textwidth]{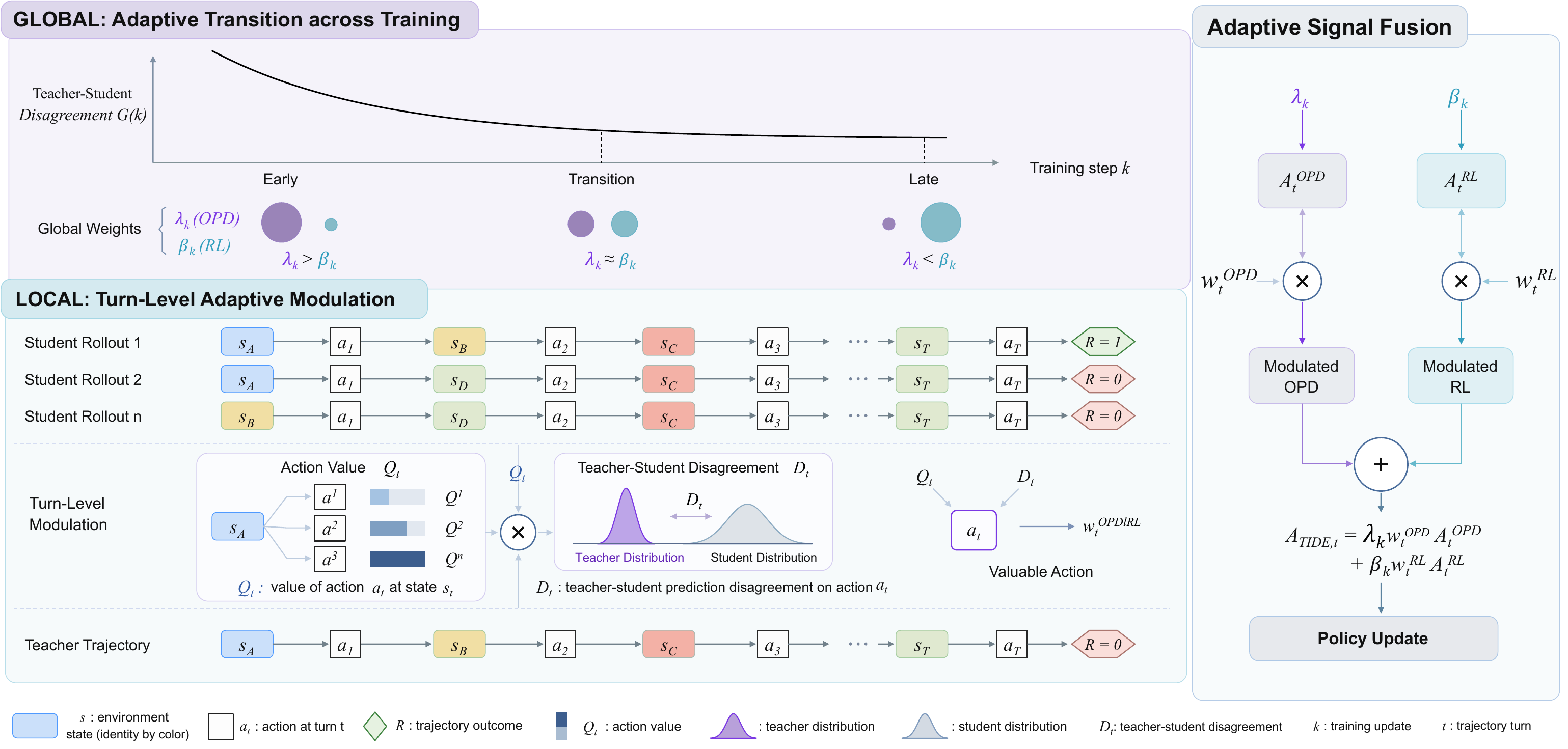}
  \caption{Overview of \tide. The global handoff shifts relative weight from OPD to RL during training, while local modulation allocates their signals across turns.}
  \label{fig:tide_overview}
\end{figure}

We address this allocation problem with \emph{Teacher--Student Transition via Informative Distillation and Exploration} (\tide), illustrated in Figure~\ref{fig:tide_overview}. Globally, \tide uses batch-level teacher--student disagreement as a practical event signal. When its positive reduction becomes slow, \tide gradually and monotonically shifts optimization weight from OPD to RL, adapting transition timing to each task's observed training dynamics without manually specifying a task-specific handoff step. Each transition increment reduces the OPD coefficient and increases the RL coefficient. Locally, \tide modulates learning signals across turns. Building on GiGPO, it derives a relative action value for each turn. Its trajectory-relative priority is combined with disagreement to prioritize teacher guidance, whereas its value is scaled by disagreement to form reward-driven updates. Combined with the global controller, these local signals allocate teacher guidance and reward-driven updates to turns according to their relative value and teacher--student disagreement.

Experiments on WebShop and ALFWorld assess \tide's two-scale design. At 1.5B, \tide achieves 77.2\% WebShop success and 86.0\% ALFWorld average success; at 3B and 7B, it attains the highest WebShop success among the evaluated baselines. Controlled ablations and analyses further support the effectiveness of the global handoff and local modulation.

Overall, our contributions are as follows:
\begin{itemize}
    \item We formulate hybrid OPD--RL coordination for multi-turn agents as a two-scale allocation problem: scheduling teacher influence across training stages and allocating update priority across turns within each trajectory.
    \item We propose \tide, which uses a disagreement-triggered schedule to set a phase-wise OPD--RL balance and combines relative action value with disagreement as a turn-priority signal.
    \item We evaluate \tide on WebShop and ALFWorld across student scales and controlled ablations, supporting its two-scale OPD--RL coordination for interactive-agent training.
\end{itemize}

\section{Method}
\label{sec:method}

\subsection{Problem Formulation}
\label{sec:problem}

\paragraph{\textbf{Multi-turn agent interaction.}}
Given a task $x\sim\mathcal{D}$ and an initial observation $o_1$, the agent repeatedly reasons, acts, and receives environment feedback. At turn $t$, its interaction history $h_t=(x,o_1,y_1,a_1,\ldots,o_t)$ contains the task and all preceding responses, executed actions, and observations. The policy $\pi_\theta$ generates a textual response $y_t\sim\pi_\theta(\cdot\mid h_t)$, consisting of reasoning followed by an environment-specific command. A task-specific parser extracts the executable action $a_t$ from $y_t$, and $\mathcal{E}$ returns the next observation $o_{t+1}$. We write the resulting turn as $u_t=(y_t,a_t,o_{t+1})$ and the complete $T$-turn rollout as $\tau=(x,o_1,u_1,u_2,\ldots,u_T)$. The interaction terminates when the agent submits a final answer, the environment signals completion, or the turn budget is exhausted.

\subsection{Preliminary Studies on OPD--RL Coordination}
\label{sec:motivation}

\paragraph{\textbf{Training-stage variation in the OPD--RL balance.}}
To examine how the relative utility of teacher supervision and reward optimization changes over training, we compare the task success of GRPO, OPD with a task-trained 7B teacher, and a fixed 1:1 weighting of OPD and RL on WebShop and ALFWorld, while tracking the average teacher--student log-probability disagreement.

\begin{figure}[t]
  \centering
  \includegraphics[width=0.92\textwidth]{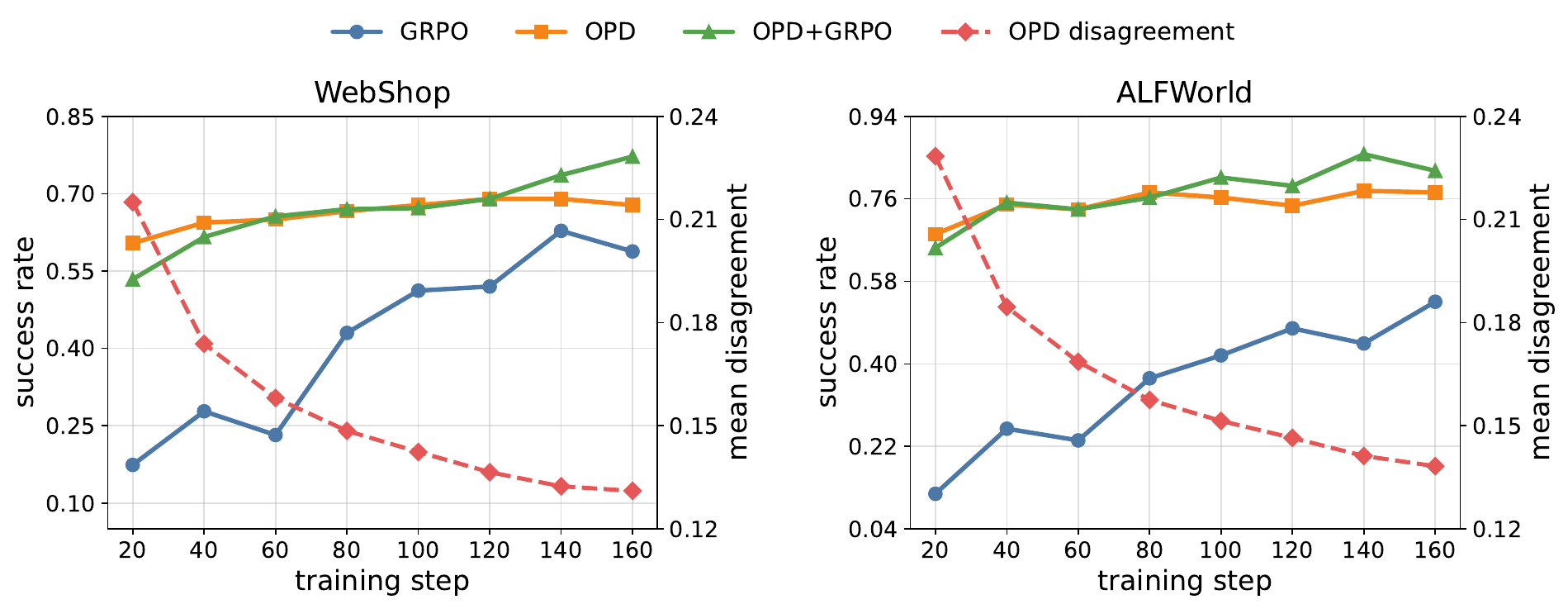}
  \caption{Training trajectories on WebShop and ALFWorld. Solid curves show task success for GRPO, OPD, and OPD+GRPO; dashed curves show teacher--student disagreement for OPD.}
  \label{fig:preexp1}
\end{figure}

In these controlled runs, OPD yields higher initial task success than GRPO on both benchmarks, but its success later plateaus or fluctuates even as teacher--student disagreement continues to fall. Combining the two objectives achieves better performance, indicating that their relative utility changes over training: stronger teacher guidance can stabilize early learning, whereas reward-driven optimization becomes more useful for later refinement. Because this transition occurs at different rates across tasks, a fixed mixing ratio cannot adapt to the changing balance.

\paragraph{\textbf{Turn-level disagreement and externally assessed process quality.}}
Teacher--student disagreement alone identifies response-level policy mismatch but does not indicate the quality of a mismatch. We analyze WebShop trajectories generated with OPD along two axes: teacher--student disagreement, computed from log-probability differences over student-sampled response tokens, and an independent GPT-5.5 process-quality score assigned to each turn. In Table~\ref{tab:preexp2_gap_quality}, $G_{\mathrm{low}}$ and $G_{\mathrm{high}}$ denote the bottom and top 20\% of turns by disagreement, and $Q_{\mathrm{low}}$ and $Q_{\mathrm{high}}$ the bottom and top 20\% by GPT-5.5 process quality. Details of the quality-scoring rubric are provided in Appendix~\ref{sec:gpt-judge}.

\begin{table}[t]
\centering
\small
\caption{Turn-level diagnostics of WebShop trajectories generated with OPD, grouped by teacher--student disagreement and GPT-5.5 process quality.}
\label{tab:preexp2_gap_quality}
\resizebox{\textwidth}{!}{
\begin{tabular}{ccccccccc}
\toprule
\multirow[c]{2}{*}[-2.5pt]{Group} & \multirow[c]{2}{*}[-2.5pt]{\shortstack[c]{Trajectory\\success (\%)}} & \multirow[c]{2}{*}[-2.5pt]{\shortstack[c]{GPT-5.5\\process quality}} & \multirow[c]{2}{*}[-2.5pt]{\shortstack[c]{Teacher--student\\disagreement}} & \multicolumn{5}{c}{Action Distribution (\%)} \\
\cmidrule(lr){5-9}
& & & & Search Items & Open Item & Select Option & Go Back & Buy Item \\
\midrule
All & 44.9 & 2.0 & 0.131 & 18.5 & 24.9 & 39.0 & 10.6 & 6.8 \\
$G_{\mathrm{low}}$ & 56.7 & 2.2 & 0.069 & 55.7 & 6.5 & 26.7 & 1.9 & 9.2 \\
$G_{\mathrm{high}}$ & 36.8 & 1.9 & 0.189 & 3.3 & 37.6 & 43.2 & 11.4 & 4.5 \\
$Q_{\mathrm{low}}$ & 0.0 & 1.0 & 0.134 & 19.9 & 26.5 & 36.0 & 15.0 & 2.3 \\
$Q_{\mathrm{high}}$ & 100.0 & 3.0 & 0.123 & 17.2 & 21.0 & 44.3 & 5.3 & 12.2 \\
\midrule
$G_{\mathrm{high}} \cap Q_{\mathrm{low}}$ & 0.0 & 1.0 & 0.189 & 3.7 & 39.6 & 40.0 & 14.2 & 2.5 \\
$G_{\mathrm{high}} \cap Q_{\mathrm{high}}$ & 100.0 & 3.0 & 0.188 & 2.2 & 28.0 & 53.8 & 8.8 & 7.1 \\
\bottomrule
\end{tabular}
}
\end{table}

In these trajectories, response-level disagreement is distributed differently across parsed action types: low-disagreement turns are dominated by routine search actions (55.7\%), whereas high-disagreement turns are concentrated on opening items and selecting options (37.6\% and 43.2\%). Within the high-disagreement subset, however, the bottom- and top-quality turns have virtually identical disagreement (0.189 and 0.188) but opposite GPT-5.5 scores and outcomes. This external diagnostic shows that response-level disagreement identifies policy mismatch but cannot determine whether a departure is exploration supported by better outcomes or low-quality policy drift.

\subsection{Teacher--Student Transition via Informative Distillation and Exploration}
\label{sec:tide}

Guided by these findings, \tide combines OPD and RL at two granularities. For token $j$ in the student-sampled response $y_t$ at interaction turn $t$, we first measure the teacher--behavior log-probability difference:
\begin{equation}
\delta_{t,j}=\log\pi_T(y_{t,j}\mid h_t,y_{t,<j})-\log\pi_{\mathrm{old}}(y_{t,j}\mid h_t,y_{t,<j}).
\label{eq:token-disagreement}
\end{equation}
We then aggregate its absolute value over the complete student-sampled response and across the update batch to obtain turn- and batch-level discrepancy:
\begin{equation}
\gap_t=\frac{1}{|y_t|}\sum_{j=1}^{|y_t|}|\delta_{t,j}|,\qquad
G_k=\frac{1}{|\mathcal{U}_k|}\sum_{t\in\mathcal{U}_k}\gap_t.
\label{eq:turn-and-batch-disagreement}
\end{equation}
Thus, $\gap_t$ is a response-level token discrepancy, including reasoning and executable-command tokens, rather than a divergence between full action distributions.

\paragraph{\textbf{Global adaptation: discrepancy-triggered handoff.}}
The global controller uses the observed discrepancy trend as an event signal for a monotone transition from OPD to RL. It determines transition timing online from the observed training dynamics of each task, without manually specifying a task-specific handoff step. To attenuate batch-level noise, we smooth $G_k$ with an exponential moving average $m_k$ and measure its relative reduction $I_k$ over a window of $W$ updates:
\begin{equation}
m_k=\mu m_{k-1}+(1-\mu)G_k,\qquad
I_k=\frac{m_{k-W+1}-m_k}{\max(m_{k-W+1},\epsilon)},
\label{eq:gap-progress}
\end{equation}
The EMA makes the event rule depend on a persistent trend rather than a single noisy batch. When $I_k$ is large, the controller leaves the handoff state unchanged; when $0<I_k\leq\zeta$, the discrepancy is decreasing slowly but remains positive, and the controller advances the RL-favored phase:
\begin{equation}
r_k=
\begin{cases}
\min(r_{k-1}+\eta,1), & 0<I_k\leq\zeta,\\
r_{k-1}, & \text{otherwise}.
\end{cases}
\label{eq:readiness}
\end{equation}
Each trigger advances the handoff state by the bounded increment $\eta$, yielding a gradual transition to RL.
The handoff state determines the weights of the two branches:
\begin{equation}
\lambda_k=\lambda_{\min}+(\lambda_{\max}-\lambda_{\min})(1-r_k),\qquad
\beta_k=\beta_{\min}+(\beta_{\max}-\beta_{\min})r_k.
\label{eq:global-coefficients}
\end{equation}
As training progresses, the observed discrepancy trend gradually decreases the OPD coefficient and increases the RL coefficient.

\paragraph{\textbf{Local adaptation: relative-value--disagreement modulation.}}
We use a relative process-reward signal, $Q_t$, to indicate whether an action leads to a better or worse outcome than matched alternatives. Within each matched-history group, its magnitude reflects the extent of this difference. Its detailed computation, following GiGPO~\citep{feng2026group}, is provided in Appendix~\ref{sec:local-details}.

Within each trajectory $\tau$, we map relative process value $Q_t$ and turn-level disagreement $\gap_t$ to comparable priorities, and combine them into an OPD turn-priority score. Let $\mathcal{V}_\tau=\{t\in\tau:\mathcal{H}_t\neq\emptyset\}$ denote turns with at least one value-discriminative matched-history group, i.e., a group with another member and nonzero return variation:
\begin{equation}
\qscore_t=\mathcal{N}_{\mathcal{V}_\tau}(Q_t),\qquad
\dscore_t=\mathcal{N}_{\tau}(\gap_t),\qquad
z_t=\qscore_t\times\dscore_t.
\label{eq:local-score}
\end{equation}
For the value signal, normalization is computed only over valid turns; let $\Delta_{\mathcal{V}_\tau}(Q)$ denote the range of $Q_t$ over $\mathcal{V}_\tau$:
\begin{equation}
\mathcal{N}_{\mathcal{V}_\tau}(Q_t)=
\begin{cases}
\dfrac{Q_t-\min_{u\in\mathcal{V}_\tau}Q_u}
{\max_{u\in\mathcal{V}_\tau}Q_u-\min_{u\in\mathcal{V}_\tau}Q_u+\epsilon}, & t\in\mathcal{V}_\tau,\ \Delta_{\mathcal{V}_\tau}(Q)>\epsilon,\\[3pt]
1, & \text{otherwise},
\end{cases}
\label{eq:trajectory-normalization}
\end{equation}
For disagreement, we use the standard min--max normalization over all turns,
\begin{equation*}
\mathcal{N}_{\tau}(\gap_t)=
\begin{cases}
\dfrac{\gap_t-\min_{u\in\tau}\gap_u}
{\max_{u\in\tau}\gap_u-\min_{u\in\tau}\gap_u+\epsilon}, & \Delta_\tau(\gap)>\epsilon,\\[3pt]
1, & \Delta_\tau(\gap)\leq\epsilon.
\end{cases}
\end{equation*}
A turn without a value-discriminative matched group retains $Q_t=0$ and uses $\qscore_t=1$ as a disagreement-only OPD fallback. The same fallback applies when valid turns have no discriminative action values. Otherwise, normalization maps the signals to non-negative priorities while preserving their ordering. We use this score for OPD, with $w^{\mathrm{OPD}}_t=z_t$. For the RL branch, normalized disagreement is the turn weight. Thus, relative action value determines the direction of reward-driven updating, while disagreement determines its magnitude:
\begin{equation}
w^{\mathrm{OPD}}_t=z_t,\qquad
w^{\mathrm{RL}}_t=\dscore_t.
\label{eq:local-weights}
\end{equation}
Consequently, high-value, high-disagreement turns receive strong positive RL updates, whereas low-value, high-disagreement turns receive strong negative RL updates. The OPD and combined advantages are defined as
\begin{equation}
\begin{aligned}
A^{\mathrm{RL}}_{t,j} &= Q_t,
\qquad A^{\mathrm{OPD}}_{t,j} = \delta_{t,j},\\
A^{\mathrm{TIDE}}_{t,j}
&=\beta_k w^{\mathrm{RL}}_t A^{\mathrm{RL}}_{t,j}
+\lambda_k w^{\mathrm{OPD}}_t A^{\mathrm{OPD}}_{t,j}.
\end{aligned}
\label{eq:tide-advantage}
\end{equation}
We optimize the advantage with the clipped surrogate objective:
\begin{equation}
\mathcal{J}(\theta)=\mathbb{E}\Bigl[\sum_{t,j}\min\!\bigl(\rho_{t,j}\,A^{\mathrm{TIDE}}_{t,j},\;\operatorname{clip}(\rho_{t,j},1{-}c,1{+}c)\,A^{\mathrm{TIDE}}_{t,j}\bigr)\Bigr],
\label{eq:tide-objective}
\end{equation}
Here, $Q_t$, $w^{\mathrm{RL}}_t$, and $w^{\mathrm{OPD}}_t$ are turn-level quantities shared by all tokens in $y_t$, whereas $\delta_{t,j}$ is token-level. The importance ratio $\rho_{t,j}$ is between the current and behavior policies:
\begin{equation}
\rho_{t,j}=\frac{\pi_\theta(y_{t,j}\mid h_t,y_{t,<j})}{\pi_{\mathrm{old}}(y_{t,j}\mid h_t,y_{t,<j})}
\label{eq:importance-ratio}
\end{equation}
The local signals allocate teacher and reward updates across turns, while the global schedule determines their phase-specific OPD--RL balance. Coupled with the global handoff, \tide provides more reliable distillation signals during the OPD-weighted early phase and promotes high-value exploration during the RL-weighted later phase.

\section{Experiments}
\label{sec:experiments}
\newcommand{\tstd}[1]{\,$\scriptstyle\pm #1$}

\subsection{Experimental Setup}
\label{sec:setup}

\paragraph{Benchmarks and metrics.}
We evaluate \tide on WebShop~\citep{yao2022webshop} and ALFWorld~\citep{shridhar2020alfworld}. On the full 500-task WebShop evaluation set, we report exact success rate and normalized task score, using up to 30 interaction steps and temperature $1.0$ with top-$p$ $1.0$ decoding. Main ALFWorld results use the official valid-seen split (140 tasks), while controlled ablations and sensitivity analyses additionally report valid-unseen results; both splits use identical task prompts, a two-step interaction history, at most 50 environment steps, temperature $0.4$, and top-$p$ $1.0$ decoding. For every method evaluated by us and for every controlled ablation, each ``$\pm$'' denotes the sample standard deviation across $n{=}3$ independently trained random seeds. Appendix~\ref{sec:searchqa-results} also reports a SearchQA experiment.

\paragraph{Models and baselines.}
We train Qwen2.5-Instruct students at the 1.5B, 3B, and 7B scales~\citep{qwen2025qwen25technicalreport}. Table~\ref{tab:controlled-main} groups the compared methods by training signal: Vanilla is a prompting baseline; GRPO and GiGPO~\citep{feng2026group} are reward-optimization baselines; OPD and OPSD are distillation references; and ATOD~\citep{tan2026atod} and SDAR are hybrid OPD--RL baselines. At every scale, OPD, ATOD, and \tide use the same frozen task-trained Qwen2.5-7B teacher and matched training configuration. Full baseline descriptions are in Appendix~\ref{sec:implementation-details}.

\paragraph{Training details.}
WebShop and ALFWorld use 16 prompts with 8 rollouts per prompt and train for 160 updates. We optimize students with AdamW at a learning rate of $10^{-6}$ and weight decay $0.01$. PPO performs one epoch per update with an importance-ratio clipping radius of $c{=}0.2$ and entropy coefficient $0.001$; minibatch sizes are 64 on WebShop and 256 on ALFWorld. Disagreement is smoothed with an exponential moving average of $\mu{=}0.9$ over a $W{=}10$-update window, and the handoff rates are $\eta{=}\zeta{=}0.02$. The OPD and RL coefficients are bounded by $\lambda\in[0.1,1.0]$ and $\beta\in[0.1,1.0]$, respectively; training therefore starts with $\lambda_0=1.0$ and $\beta_0=0.1$. Detailed relative-action-value configurations are provided in Appendix~\ref{sec:local-details}.

\begin{table*}[!t]
\caption{Main results across model scales on ALFWorld and WebShop (\%). Prior-work values are marked \textsuperscript{\ensuremath{\dagger}}. Bold indicates the highest displayed value in each column at each scale.}
\label{tab:controlled-main}
\centering
\scriptsize
\setlength{\tabcolsep}{3.2pt}
\renewcommand{\arraystretch}{0.92}
\resizebox{\textwidth}{!}{%
\begin{tabular}{llccccccccc}
\toprule
\multirow{2}{*}{\textbf{Method}} & \multirow{2}{*}{\textbf{Type}} & \multicolumn{7}{c}{\textbf{ALFWorld}} & \multicolumn{2}{c}{\textbf{WebShop}} \\
\cmidrule(lr){3-9}\cmidrule(lr){10-11}
& & \textbf{Pick} & \textbf{Look} & \textbf{Clean} & \textbf{Heat} & \textbf{Cool} & \textbf{Pick2} & \textbf{Avg.} & \textbf{Score} & \textbf{SR} \\
\midrule
\rowcolor{gray!12}\multicolumn{11}{l}{\textit{Qwen2.5-1.5B-Instruct}} \\
\midrule
Vanilla\textsuperscript{\ensuremath{\dagger}} & Prompt & 11.1 & 0.0 & 6.2 & 0.0 & 0.0 & 4.2 & 5.5 & 17.8 & 5.5 \\
GRPO & RL & 80.0\tstd{2.9} & 53.8\tstd{7.7} & 50.6\tstd{4.3} & 37.5\tstd{6.3} & 68.0\tstd{4.0} & 44.4\tstd{4.8} & 58.8\tstd{0.8} & 82.4\tstd{1.2} & 62.8\tstd{2.0} \\
GiGPO\textsuperscript{\ensuremath{\dagger}} & RL & 94.4 & 67.5 & 94.8 & \textbf{94.4} & 79.8 & \textbf{76.4} & \textbf{86.7} & 83.1 & 65.0 \\
OPD & Distill & 88.6\tstd{5.0} & 66.7\tstd{4.4} & 84.0\tstd{4.0} & 72.9\tstd{5.1} & 61.3\tstd{4.7} & 56.9\tstd{5.5} & 73.6\tstd{2.1} & 79.4\tstd{2.3} & 67.8\tstd{3.1} \\
OPSD\textsuperscript{\ensuremath{\dagger}} & Distill & 26.3 & 16.7 & 9.1 & 6.7 & 9.1 & 5.3 & 14.1 & 22.3 & 10.2 \\
ATOD & Hybrid & 88.6\tstd{5.7} & \textbf{76.9}\tstd{7.7} & \textbf{96.3}\tstd{3.7} & 93.8\tstd{4.1} & 80.0\tstd{4.0} & 62.5\tstd{4.2} & 83.6\tstd{3.0} & 85.5\tstd{1.1} & 73.2\tstd{2.4} \\
\textbf{\tide} & Hybrid & \textbf{95.2}\tstd{3.3} & 71.8\tstd{4.4} & 88.9\tstd{3.7} & 93.8\tstd{6.3} & \textbf{85.3}\tstd{2.3} & 72.2\tstd{4.8} & 86.0\tstd{0.8} & \textbf{89.8}\tstd{1.2} & \textbf{77.2}\tstd{1.8} \\
\midrule
\rowcolor{gray!12}\multicolumn{11}{l}{\textit{Qwen2.5-3B-Instruct}} \\
\midrule
Vanilla\textsuperscript{\ensuremath{\dagger}} & Prompt & 44.4 & 11.1 & 6.2 & 15.4 & 28.6 & 12.5 & 21.9 & 6.7 & 0.8 \\
GRPO & RL & 91.4\tstd{2.9} & 61.5\tstd{7.7} & 96.3\tstd{3.7} & 62.5\tstd{6.3} & 65.3\tstd{4.6} & 47.2\tstd{4.8} & 74.0\tstd{0.8} & 79.8\tstd{1.1} & 63.3\tstd{2.0} \\
OPD & Distill & 90.5\tstd{3.4} & 74.4\tstd{4.4} & 90.1\tstd{3.3} & 77.1\tstd{4.1} & 61.3\tstd{4.4} & 52.8\tstd{5.0} & 75.7\tstd{2.0} & 80.1\tstd{2.1} & 69.3\tstd{2.9} \\
OPSD\textsuperscript{\ensuremath{\dagger}} & Distill & 48.8 & 41.7 & 16.7 & 0.0 & 15.8 & 16.7 & 28.1 & 11.3 & 3.1 \\
SDAR\textsuperscript{\ensuremath{\dagger}} & Hybrid & \textbf{97.1} & 62.5 & \textbf{100.0} & 61.9 & 75.0 & \textbf{84.2} & 84.4 & 85.0 & 68.0 \\
ATOD & Hybrid & 94.3\tstd{4.9} & \textbf{76.9}\tstd{4.4} & 96.3\tstd{5.4} & 87.5\tstd{6.3} & 88.0\tstd{3.9} & 66.7\tstd{4.2} & 86.4\tstd{2.6} & 86.6\tstd{0.8} & 74.1\tstd{2.0} \\
\textbf{\tide} & Hybrid & 94.3\tstd{2.9} & \textbf{76.9}\tstd{7.7} & 92.6\tstd{3.7} & \textbf{93.8}\tstd{6.3} & \textbf{96.0}\tstd{4.0} & 75.0\tstd{4.2} & \textbf{89.3}\tstd{0.7} & \textbf{90.2}\tstd{0.9} & \textbf{79.0}\tstd{1.5} \\
\midrule
\rowcolor{gray!12}\multicolumn{11}{l}{\textit{Qwen2.5-7B-Instruct}} \\
\midrule
Vanilla\textsuperscript{\ensuremath{\dagger}} & Prompt & 36.1 & 22.2 & 3.1 & 0.0 & 0.0 & 0.0 & 12.5 & 5.9 & 1.6 \\
GRPO & RL & 91.4\tstd{2.9} & 87.2\tstd{4.4} & 96.3\tstd{3.7} & 81.3\tstd{6.3} & 65.3\tstd{4.6} & 58.3\tstd{4.2} & 80.5\tstd{0.8} & 80.9\tstd{1.0} & 72.6\tstd{1.7} \\
GiGPO\textsuperscript{\ensuremath{\dagger}} & RL & \textbf{97.7} & 82.7 & 98.8 & 83.7 & 89.3 & \textbf{79.2} & 90.8 & 84.4 & 72.8 \\
OPD & Distill & 91.4\tstd{2.8} & 84.6\tstd{4.0} & 95.1\tstd{3.0} & 79.2\tstd{3.8} & 64.0\tstd{4.1} & 56.9\tstd{4.6} & 79.3\tstd{1.9} & 80.5\tstd{1.8} & 71.7\tstd{2.6} \\
OPSD\textsuperscript{\ensuremath{\dagger}} & Distill & 50.0 & 60.0 & 22.7 & 21.4 & 17.6 & 9.5 & 32.8 & 4.5 & 2.3 \\
SDAR\textsuperscript{\ensuremath{\dagger}} & Hybrid & 94.7 & 75.0 & \textbf{100.0} & 86.7 & 68.2 & 78.9 & 85.9 & 89.4 & 82.8 \\
ATOD & Hybrid & 97.1\tstd{2.9} & 84.6\tstd{7.7} & 96.3\tstd{3.7} & 93.8\tstd{2.1} & 88.0\tstd{3.2} & 70.8\tstd{3.3} & 89.3\tstd{2.1} & 89.1\tstd{0.7} & 79.0\tstd{1.5} \\
\textbf{\tide} & Hybrid & 95.2\tstd{1.6} & \textbf{92.3}\tstd{7.7} & 95.1\tstd{2.1} & \textbf{97.9}\tstd{3.6} & \textbf{94.7}\tstd{2.3} & 77.8\tstd{2.4} & \textbf{92.1}\tstd{0.7} & \textbf{92.1}\tstd{0.7} & \textbf{83.2}\tstd{1.3} \\
\bottomrule
\end{tabular}
}
\end{table*}

\subsection{Main Results}
\label{sec:results}

Table~\ref{tab:controlled-main} compares \tide with prompting, reward-optimization, distillation, and hybrid baselines across Qwen2.5-1.5B, 3B, and 7B scales. Values marked with \textsuperscript{\ensuremath{\dagger}} are reported from prior work; every unmarked row was evaluated by us under the matched protocol. Prompting baselines remain weak on both benchmarks. Among reward-optimization methods, GiGPO reports the highest displayed ALFWorld average at 1.5B, while \tide achieves the highest displayed WebShop success rate at all three scales. Among distillation methods, task-trained OPD consistently outperforms the OPSD reference. For hybrid methods, the controlled ATOD--\tide comparison holds the teacher and training setting fixed: \tide exceeds ATOD on both benchmarks at every scale. At 7B, \tide also exceeds the displayed SDAR reference on WebShop success (83.2\% vs.\ 82.8\%) and ALFWorld average success (92.1\% vs.\ 85.9\%), and achieves the highest displayed ALFWorld average (92.1\%).

\subsection{Ablation Studies}
\label{sec:ablations}

\subsubsection{Discrepancy-triggered handoff against alternative schedules}

Table~\ref{tab:ablation}(a) tests whether using the measured discrepancy trend as a schedule signal improves on fixed, preset, and success-guided alternatives in our evaluated setting. All hybrid variants use the same Qwen2.5-1.5B-Instruct student and frozen GRPO-trained Qwen2.5-7B-Instruct teacher; they differ only in the evolution of the global coefficients. We compare \tide with GRPO, Success Handoff, Fixed 1:1 Mixture, and preset linear and cosine handoffs. Definitions of the time schedules, together with realized handoff trajectories and parameter sensitivity, are provided in Appendix~\ref{sec:per-task-handoff}.

\tide's discrepancy-triggered handoff outperforms all fixed, preset, and success-guided alternatives. Success Handoff reaches 71.4\% WebShop success and 73.6\%/73.1\% ALFWorld seen/unseen success, below the fixed and preset schedules. Fixed 1:1 Mixture improves over GRPO on both benchmarks, confirming the value of combining the two signals. Among preset schedules, Cosine Handoff is strongest on WebShop at 73.2\% success; \tide exceeds it by 4.0 points on WebShop success and 3.8 points on ALFWorld unseen success.

\begin{table}[t]
\caption{TIDE ablations with Qwen2.5-1.5B-Instruct students (\%). Bold indicates the highest displayed value.}
\label{tab:ablation}
\centering
\begin{minipage}[t]{0.485\textwidth}\centering
(a) Global handoff strategies\\[4pt]
\scriptsize\setlength{\tabcolsep}{1.8pt}\renewcommand{\arraystretch}{1.10}
\begin{tabular*}{\linewidth}{@{\extracolsep{\fill}}>{\centering\arraybackslash}m{0.30\linewidth}cccc@{}}
\toprule
\multirow{2}{*}{\raisebox{-0.5ex}{\textbf{Method}}} & \multicolumn{2}{c}{\textbf{WebShop}} & \multicolumn{2}{c}{\textbf{ALFWorld}} \\
\cmidrule(lr){2-3}\cmidrule(lr){4-5}
& \mbox{\textbf{SR $\uparrow$}} & \mbox{\textbf{Score $\uparrow$}} & \mbox{\textbf{Seen $\uparrow$}} & \mbox{\textbf{Unseen $\uparrow$}} \\
\midrule
GRPO & 62.8\tstd{2.0} & 82.4\tstd{1.2} & 58.8\tstd{0.8} & 47.8\tstd{2.6} \\
Success Handoff & 71.4\tstd{2.5} & 82.2\tstd{1.6} & 73.6\tstd{2.5} & 73.1\tstd{2.7} \\
Fixed 1:1 Mixture & 72.0\tstd{1.7} & 84.4\tstd{0.8} & 77.1\tstd{1.4} & 79.9\tstd{1.5} \\
Linear Handoff & 72.2\tstd{2.2} & 86.3\tstd{1.3} & 85.7\tstd{1.9} & 81.3\tstd{2.0} \\
Cosine Handoff & 73.2\tstd{1.4} & 87.3\tstd{0.7} & 85.0\tstd{1.4} & 81.3\tstd{1.5} \\
\tide & \textbf{77.2}\tstd{1.8} & \textbf{89.8}\tstd{1.2} & \textbf{86.0}\tstd{0.8} & \textbf{85.1}\tstd{1.5} \\
\bottomrule\end{tabular*}\end{minipage}\hfill
\begin{minipage}[t]{0.485\textwidth}\centering
(b) Local signal allocation\\[4pt]
\scriptsize\setlength{\tabcolsep}{1.8pt}\renewcommand{\arraystretch}{1.10}
\begin{tabular*}{\linewidth}{@{\extracolsep{\fill}}>{\centering\arraybackslash}m{0.30\linewidth}cccc@{}}
\toprule
\multirow{2}{*}{\raisebox{-0.5ex}{\textbf{Method}}} & \multicolumn{2}{c}{\textbf{WebShop}} & \multicolumn{2}{c}{\textbf{ALFWorld}} \\
\cmidrule(lr){2-3}\cmidrule(lr){4-5}
& \mbox{\textbf{SR $\uparrow$}} & \mbox{\textbf{Score $\uparrow$}} & \mbox{\textbf{Seen $\uparrow$}} & \mbox{\textbf{Unseen $\uparrow$}} \\
\midrule
w/o OPD Mod. & 74.4\tstd{1.5} & 85.5\tstd{0.8} & 83.6\tstd{1.2} & 80.6\tstd{1.5} \\
w/o RL Mod. & 72.6\tstd{2.4} & 84.7\tstd{1.4} & 85.0\tstd{1.9} & 79.1\tstd{2.0} \\
w/o Process Reward & 72.2\tstd{2.6} & 84.4\tstd{1.7} & 80.0\tstd{2.1} & 79.1\tstd{2.2} \\
w/o Disagreement & 74.4\tstd{1.6} & 85.0\tstd{1.0} & 83.6\tstd{1.4} & 79.9\tstd{1.5} \\
Additive Fusion & 67.4\tstd{3.8} & 74.7\tstd{2.7} & 70.7\tstd{2.5} & 70.1\tstd{2.2} \\
\tide & \textbf{77.2}\tstd{1.8} & \textbf{89.8}\tstd{1.2} & \textbf{86.0}\tstd{0.8} & \textbf{85.1}\tstd{1.5} \\
\bottomrule\end{tabular*}\end{minipage}
\end{table}

\subsubsection{Relative-value--disagreement ablation}
Following the setup in Section~\ref{sec:ablations}, Table~\ref{tab:ablation}(b) evaluates the design of TIDE's local allocation. Every variant retains TIDE's global handoff, teacher, training budget, and OPD--RL objective, and differs only in its local allocation. We compare removing the process-reward or disagreement signal, replacing their product with Additive Fusion, and disabling local modulation in the OPD or RL branch. Detailed definitions of these variants are provided in Appendix~\ref{sec:local-details}.

The full method reaches 77.2\% WebShop SR and outperforms every local-allocation variant. Removing the process-reward or disagreement signal reduces SR to 72.2\% or 74.4\%, while Additive Fusion reduces it to 67.4\%, showing that both signals and their multiplicative fusion are necessary. Disabling local modulation in the RL or OPD branch reduces SR to 72.6\% or 74.4\%. Thus, the strongest performance requires joint local modulation of both OPD and RL branches.

\subsection{Analysis}
\label{sec:analysis}

\subsubsection{Global handoff dynamics}

\begin{figure}[t]
  \centering
  \begin{subfigure}[t]{0.325\textwidth}
    \centering
    \includegraphics[width=\linewidth]{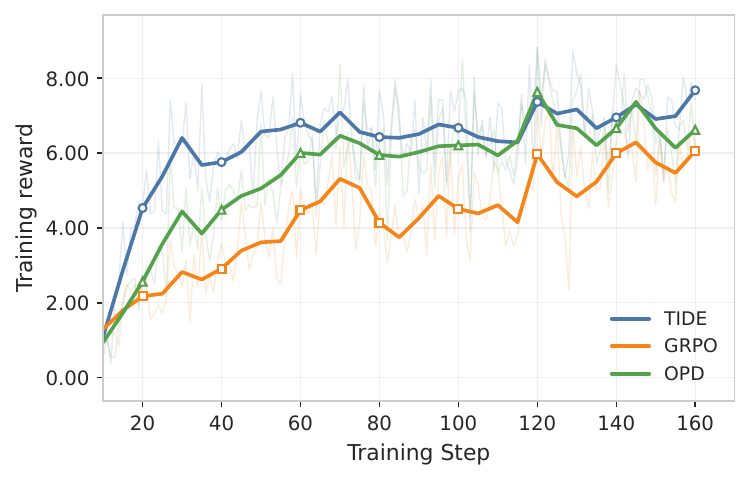}
    \caption{Training reward.}
    \label{fig:analysis1-reward}
  \end{subfigure}\hfill
  \begin{subfigure}[t]{0.325\textwidth}
    \centering
    \includegraphics[width=\linewidth]{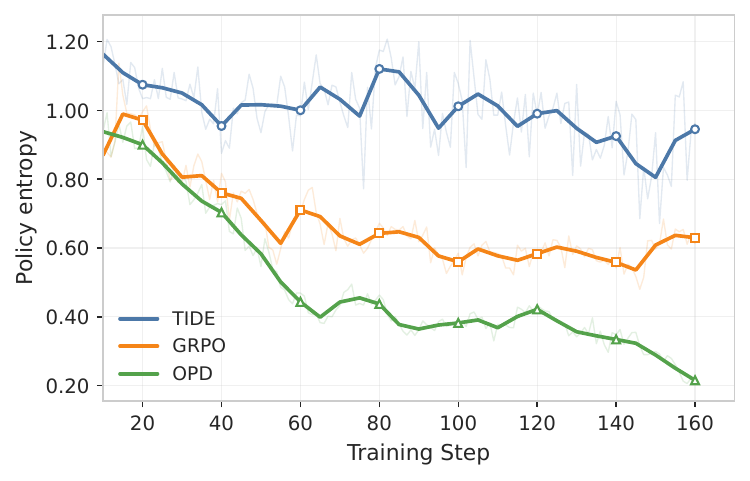}
    \caption{Policy entropy.}
    \label{fig:analysis1-entropy}
  \end{subfigure}\hfill
  \begin{subfigure}[t]{0.325\textwidth}
    \centering
    \includegraphics[width=\linewidth]{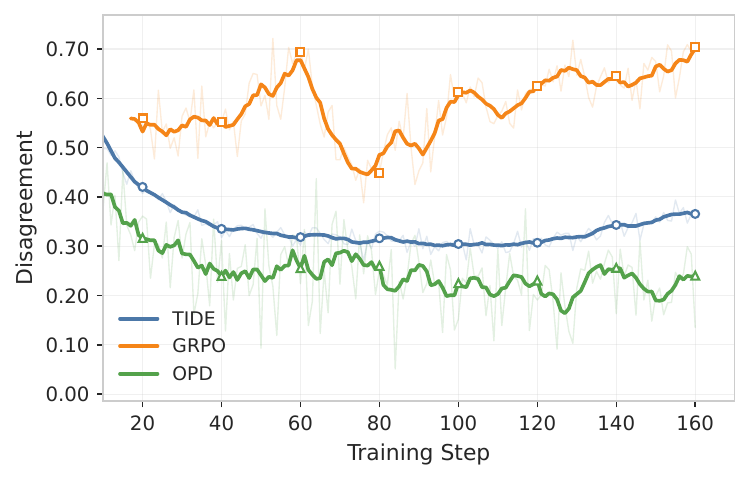}
    \caption{Teacher--student disagreement.}
    \label{fig:analysis1-gap}
  \end{subfigure}
  \caption{Training dynamics of TIDE and baselines on WebShop. Faint lines show per-step values; bold curves show smoothed trends.}
  \label{fig:analysis1}
\end{figure}

To characterize how the discrepancy-triggered handoff changes the optimization, we track training reward, policy entropy, and teacher--student disagreement throughout training. All methods use the 1.5B setup in Section~\ref{sec:setup}; \tide and OPD share the frozen GRPO-trained 7B teacher. Appendix~\ref{sec:per-task-handoff} additionally reports the realized benchmark-level handoff trajectories and parameter sensitivity.

Figure~\ref{fig:analysis1} shows that \tide reaches higher training reward earlier than GRPO and OPD on this WebShop setting while maintaining higher policy entropy. Its teacher--student disagreement decreases rapidly in the early updates and later stabilizes at an intermediate level between GRPO and OPD. Together, these trajectories illustrate the intended progression: teacher-weighted updates stabilize learning early, while the later increase in reward optimization preserves policy diversity and supports refinement toward higher-value behavior. This plot characterizes the resulting training dynamics; the disagreement statistic is a practical event signal for setting the schedule, not a direct measurement of teacher utility or RL reliability.

\subsubsection{Turn-level learning-signal allocation}
\label{sec:analysis-local}

To understand how local modulation shapes behavior, we examine turn-level OPD signal strength across parsed agent-action types under relative-value-only, disagreement-only, and joint modulation. We group turns from the same \tide rollouts by action type and normalize the mean local weight of each group by the mean over all turns. Thus, a value above or below one indicates that turns of the corresponding type are emphasized or suppressed. The underlying disagreement remains the response-level statistic in Equation~\ref{eq:turn-and-batch-disagreement}. We use the same 1.5B WebShop configuration as Section~\ref{sec:setup}, including the rollout budget and optimization settings.

\begin{figure}[t]
  \centering
  \begin{subfigure}[t]{0.325\textwidth}
    \centering
    \includegraphics[width=\linewidth]{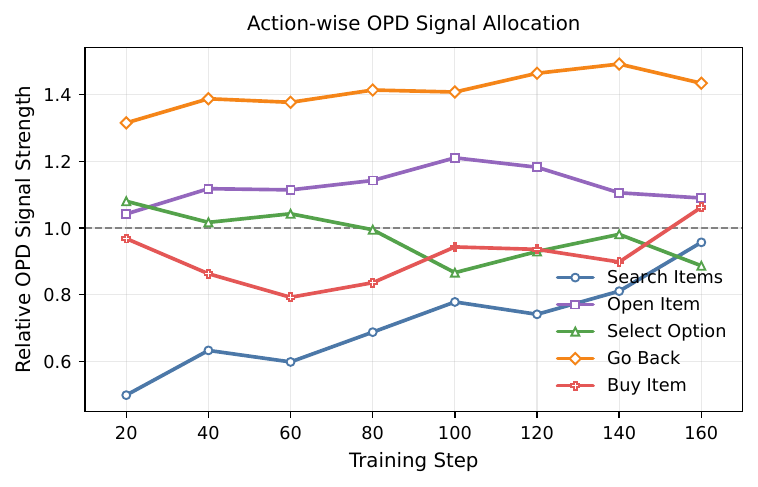}
    \caption{Relative-value diagnostic.}
    \label{fig:analysis2-value}
  \end{subfigure}\hfill
  \begin{subfigure}[t]{0.325\textwidth}
    \centering
    \includegraphics[width=\linewidth]{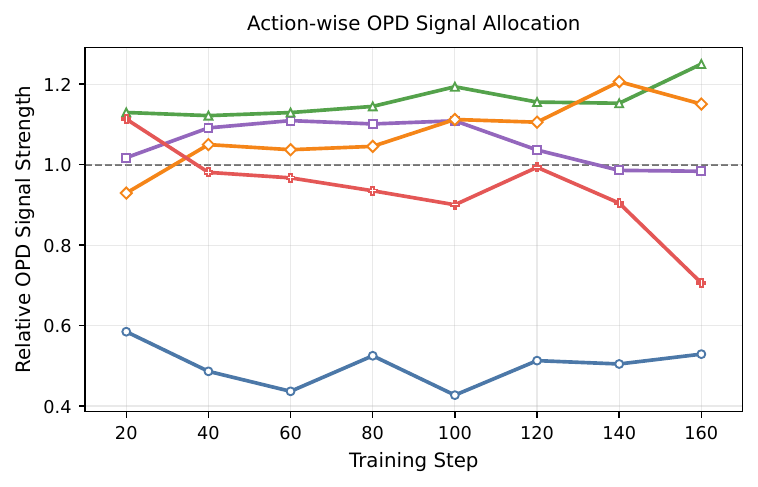}
    \caption{Disagreement diagnostic.}
    \label{fig:analysis2-disagreement}
  \end{subfigure}\hfill
  \begin{subfigure}[t]{0.325\textwidth}
    \centering
    \includegraphics[width=\linewidth]{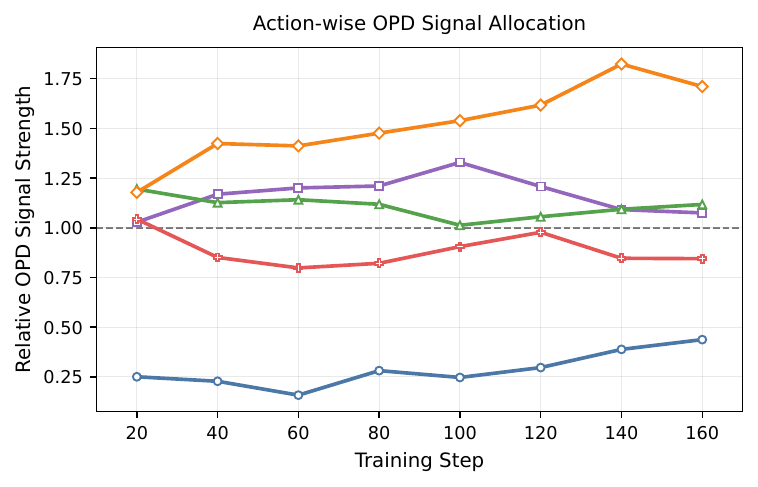}
    \caption{\tide OPD allocation.}
    \label{fig:analysis2-opd-joint}
  \end{subfigure}
  \caption{Turn-level learning-signal allocation by parsed action type. Relative-value and disagreement diagnostics, followed by \tide's joint OPD allocation.}
  \label{fig:analysis2}
\end{figure}

Figure~\ref{fig:analysis2} highlights Go Back, an action that returns to a previous page after an unproductive search or option-selection branch. Such corrective actions can redirect the subsequent trajectory and therefore affect task completion. Disagreement-only modulation also emphasizes other teacher--student differences, whereas joint modulation assigns Go Back a higher relative OPD weight by combining disagreement with relative process value. The resulting allocation focuses teacher guidance on corrective decisions associated with favorable outcome evidence.

\subsubsection{Computational cost}

Given a frozen task-trained teacher, we measure student update time for 1.5B and 3B models on WebShop and ALFWorld, and compare WebShop episode lengths of \tide, GRPO, and OPD.

\begin{figure}[!t]
  \centering
  \begin{subfigure}[t]{0.42\textwidth}
    \centering
    \includegraphics[width=\linewidth]{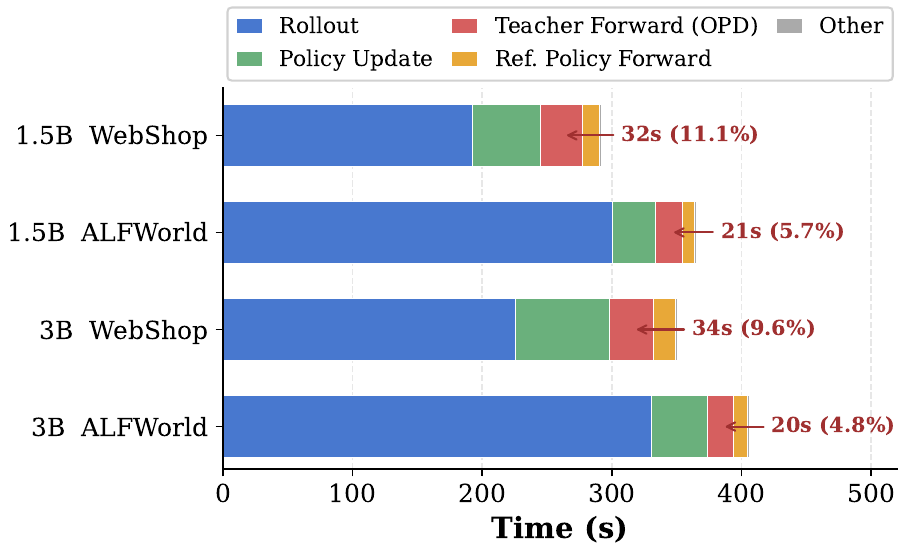}
    \caption{Student-training time per update.}
    \label{fig:overhead}
  \end{subfigure}\quad
  \begin{subfigure}[t]{0.42\textwidth}
    \centering
    \includegraphics[width=\linewidth]{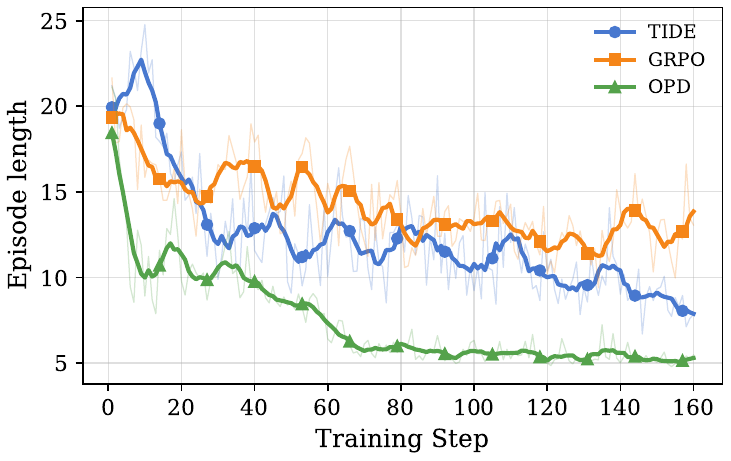}
    \caption{Interactions per episode.}
    \label{fig:analysis1-episode-length}
  \end{subfigure}
  \caption{Student-training and interaction cost of \tide, given a frozen task-trained teacher. (a) Time per student update. (b) Environment interactions per episode.}
  \label{fig:cost}
\end{figure}

Figure~\ref{fig:overhead} shows that rollout generation dominates update time; the teacher forward pass shared with OPD accounts for 5--11\%, and \tide-specific computation adds negligible overhead. This excludes the one-time GRPO training used to construct a teacher, which may be reused across students on the same task. Figure~\ref{fig:analysis1-episode-length} shows that, after initially longer OPD-dominated episodes, \tide uses fewer interactions than GRPO later in training while remaining above OPD. It therefore improves WebShop success while retaining efficient interaction.

\section{Related Work}
\label{sec:related}

\paragraph{\textbf{On-policy distillation.}}
On-policy distillation transfers teacher knowledge using samples drawn from the current student policy. Conventional knowledge distillation trains a compact student on teacher predictions over a fixed dataset~\citep{hinton2015distilling}; Generalized Knowledge Distillation instead scores student-generated sequences with the teacher to reduce distribution mismatch~\mbox{\citep{agarwal2024policy}}. Subsequent work constructs reasoning supervision from verified on-policy solutions~\citep{zhao2026self} or incorporates self-generated distillation targets into reward-based training~\citep{hubotter2026reinforcement,yang2026self}. For interactive agents, recent methods use extracted skills~\citep{wang2026skill}, on-policy experience~\citep{yang2026opid}, or iterative distillation~\citep{wu2026seed} to support agent self-evolution. On-policy samples are particularly important in interactive settings, where each action changes the subsequent context on which teacher guidance is evaluated. These approaches establish teacher guidance on the student's evolving trajectory distribution; \tide further adapts its allocation across both training stages and interaction turns.

\paragraph{\textbf{Credit assignment for interactive agents.}}
Fine-grained credit assignment uses intermediate evidence to identify decisions that contribute to an eventual outcome, especially when supervision is available only at the end of a long interaction. In mathematical reasoning, step-level labels~\citep{wang2024math}, divide-and-conquer search~\citep{luo2024improve}, implicit process rewards~\citep{cui2025process}, and rollout-based value estimation~\citep{kazemnejad2024vineppo} provide process supervision. For interactive agents, GiGPO uses repeated environment states to construct relative process advantages~\citep{feng2026group}, HGPO uses matched observation histories to estimate hierarchical process credit~\citep{he2026hierarchy}, and GraphGPO propagates credit through graph-structured interaction histories~\citep{cheng2026beyond}. These methods provide outcome-relevant evidence at a finer granularity than trajectory-level rewards. Building on this line, \tide uses a history-based relative process-value signal as outcome evidence for local OPD--RL allocation and combines it with teacher--student disagreement to prioritize turns for both branches.

\paragraph{\textbf{Hybrid distillation and reinforcement learning.}}
Recent concurrent work studies different mechanisms for coordinating dense distillation with outcome-based policy optimization. ATOD uses a prescribed linear annealing schedule from OPD toward RL~\citep{tan2026atod}; RetireOPD retires teacher supervision once discrepancy plateaus and the student reaches a teacher-relative performance threshold~\citep{yu2026retireopd}. Other methods use signal-calibrated weighting~\citep{zheng2026scope}, sample routing~\citep{li2026unifying}, verifiable rewards~\citep{lin2026policy}, teacher distributions~\citep{liu2026teacher,wang2026distilled}, trajectory-relative normalization~\citep{zheng2026trajectory}, objective-influence estimates~\citep{lan2026trust}, or lookahead agreement~\citep{qu2026turnsight}; SDAR uses per-token gating~\citep{lu2026self}, and two-stage approaches sequence distillation and RL~\citep{ye2026opdsearch+,li2026sequential,kim2026opsd}. Together, these approaches instantiate the trade-off through schedules, routing, or local gating. \tide complements them with a shared two-scale design that coordinates the OPD--RL balance across training and update allocation across turns, using outcome-aware priority to connect the two levels.

\section{Conclusion}
We introduced \tide, a two-scale approach for coordinating teacher supervision and reward optimization in multi-turn agents. Its global controller uses a smoothed discrepancy trend to transition from OPD to RL without prescribing a task-specific handoff step. Locally, relative action value and response-level discrepancy allocate teacher-guided and reward-driven updates across turns. Results on WebShop and ALFWorld, across the evaluated student scales, show improvements over fixed, scheduled, and component alternatives.

\section*{AI Assistance Disclosure}

Large language model tools were used for language editing and organization of the manuscript, literature retrieval and citation discovery, conceptual discussion, and assistance with code-review and research-execution workflows. GPT-5.5 was additionally used only as the external turn-quality evaluator described in Appendix~\ref{sec:gpt-judge}; it did not participate in policy training, reward construction, checkpoint or hyperparameter selection, or test-time action selection. The authors reviewed all AI-assisted material and take responsibility for the final manuscript, experiments, and claims.

\bibliographystyle{iclr2027_conference}
\bibliography{iclr2027_conference}

@article{hinton2015distilling,
  title={Distilling the knowledge in a neural network},
  author={Hinton, Geoffrey and Vinyals, Oriol and Dean, Jeff},
  journal={arXiv preprint arXiv:1503.02531},
  year={2015}
}

@inproceedings{agarwal2024policy,
  title={On-policy distillation of language models: Learning from self-generated mistakes},
  author={Agarwal, Rishabh and Vieillard, Nino and Zhou, Yongchao and Stanczyk, Piotr and Ramos Garea, Sabela and Geist, Matthieu and Bachem, Olivier},
  booktitle={International Conference on Learning Representations},
  volume={2024},
  pages={21246--21263},
  year={2024}
}

@article{yao2022webshop,
  title={Webshop: Towards scalable real-world web interaction with grounded language agents},
  author={Yao, Shunyu and Chen, Howard and Yang, John and Narasimhan, Karthik},
  journal={Advances in Neural Information Processing Systems},
  volume={35},
  pages={20744--20757},
  year={2022}
}

@article{shridhar2020alfworld,
  title={Alfworld: Aligning text and embodied environments for interactive learning},
  author={Shridhar, Mohit and Yuan, Xingdi and C{\^o}t{\'e}, Marc-Alexandre and Bisk, Yonatan and Trischler, Adam and Hausknecht, Matthew},
  journal={arXiv preprint arXiv:2010.03768},
  year={2020}
}

@article{shao2024deepseekmath,
  title={Deepseekmath: Pushing the limits of mathematical reasoning in open language models},
  author={Shao, Zhihong and Wang, Peiyi and Zhu, Qihao and Xu, Runxin and Song, Junxiao and Bi, Xiao and Zhang, Haowei and Zhang, Mingchuan and Li, YK and Wu, Yang and others},
  journal={arXiv preprint arXiv:2402.03300},
  year={2024}
}

@misc{qwen2025qwen25technicalreport,
      title={Qwen2.5 Technical Report}, 
      author={Qwen and : and An Yang and Baosong Yang and Beichen Zhang and Binyuan Hui and Bo Zheng and Bowen Yu and Chengyuan Li and Dayiheng Liu and Fei Huang and Haoran Wei and Huan Lin and Jian Yang and Jianhong Tu and Jianwei Zhang and Jianxin Yang and Jiaxi Yang and Jingren Zhou and Junyang Lin and Kai Dang and Keming Lu and Keqin Bao and Kexin Yang and Le Yu and Mei Li and Mingfeng Xue and Pei Zhang and Qin Zhu and Rui Men and Runji Lin and Tianhao Li and Tianyi Tang and Tingyu Xia and Xingzhang Ren and Xuancheng Ren and Yang Fan and Yang Su and Yichang Zhang and Yu Wan and Yuqiong Liu and Zeyu Cui and Zhenru Zhang and Zihan Qiu},
      year={2025},
      eprint={2412.15115},
      archivePrefix={arXiv},
      primaryClass={cs.CL},
      url={https://arxiv.org/abs/2412.15115}, 
}

@inproceedings{he2026hierarchy,
  title={Hierarchy-of-groups policy optimization for long-horizon agentic tasks},
  author={He, Shuo and Feng, Lang and Cheng, Xin and Feng, Lei and An, Bo and others},
  booktitle={International Conference on Learning Representations},
  volume={2026},
  pages={27572--27593},
  year={2026}
}

@article{tan2026atod,
  title={ATOD: Annealed Turn-aware On-policy Distillation for Multi-turn Autonomous Agents},
  author={Tan, Qitai and Zong, Zefang and Li, Yang and Chen, Peng},
  journal={arXiv preprint arXiv:2606.27814},
  year={2026}
}

@article{zheng2026scope,
  title={Scope: Signal-calibrated on-policy distillation enhancement with dual-path adaptive weighting},
  author={Zheng, Binbin and Ma, Xing and Liang, Yiheng and Ruan, Jingqing and Fu, Xiaoliang and Lin, Kepeng and Zhu, Benchang and Zeng, Ke and Cai, Xunliang},
  journal={arXiv preprint arXiv:2604.10688},
  year={2026}
}

@article{li2026unifying,
  title={Unifying group-relative and self-distillation policy optimization via sample routing},
  author={Li, Gengsheng and Yang, Tianyu and Fang, Junfeng and Song, Mingyang and Zheng, Mao and Guo, Haiyun and Zhang, Dan and Wang, Jinqiao and Chua, Tat-Seng},
  journal={arXiv preprint arXiv:2604.02288},
  year={2026}
}

@article{lin2026policy,
  title={On-policy Distillation with Verifiable Reward},
  author={Lin, Wenze and Zhao, Jiale and Jiang, Xitai and Rao, Songde and Li, Yining and Wang, Shenzhi and He, Bingxiang and Huang, Gao},
  journal={arXiv preprint arXiv:2608.24696},
  year={2026}
}

@article{liu2026teacher,
  title={Teacher-guided policy optimization for on-policy reasoning distillation under large policy divergence},
  author={Liu, Xinyu and Jiao, Kechen and Xiao, Chunyang and Zhao, Runsong and Ruan, Junhao and Li, Bei and Liu, Jiahao and Wang, Qifan and Chen, Xin and Wang, Jingang and others},
  journal={arXiv preprint arXiv:2605.13230},
  year={2026}
}

@article{wang2026distilled,
  title={Distilled Reinforcement Learning for LLM Post-training},
  author={Wang, Chen and Li, Zhaochun and Bai, Jionghao and Zhang, Yining and Deng, Hexuan and Lan, Ge and Wang, Yue},
  journal={arXiv preprint arXiv:2607.17247},
  year={2026}
}

@article{yu2026retireopd,
  title={RetireOPD: Self-Retiring On-Policy Distillation for Agentic Reinforcement Learning},
  author={Yu, Yan and Lu, Zhengxi and Liu, Yizhou and Pan, Yichen and Wang, Aozhe and Chen, Qipeng and Yang, Hua and Zhang, Wenqi and Lu, Weiming and Chen, Qianglong and others},
  journal={arXiv preprint arXiv:2609.20784},
  year={2026}
}

@article{feng2026group,
  title={Group-in-group policy optimization for llm agent training},
  author={Feng, Lang and Xue, Zhenghai and Liu, Tingcong and An, Bo},
  journal={Advances in Neural Information Processing Systems},
  volume={38},
  pages={46375--46408},
  year={2026}
}

@article{cheng2026beyond,
  title={Beyond Trajectory-Level Attribution: Graph-Based Credit Assignment for Agentic Reinforcement Learning},
  author={Cheng, Xin and He, Shuo and Feng, Lang and Xu, HaiYang and Yan, Ming and Feng, Lei and An, Bo},
  journal={arXiv preprint arXiv:2605.26684},
  year={2026}
}

@article{zhao2026self,
  title={Self-distilled reasoner: On-policy self-distillation for large language models},
  author={Zhao, Siyan and Xie, Zhihui and Liu, Mengchen and Huang, Jing and Pang, Guan and Chen, Feiyu and Grover, Aditya},
  journal={arXiv preprint arXiv:2601.18734},
  year={2026}
}

@article{hubotter2026reinforcement,
  title={Reinforcement learning via self-distillation},
  author={H{\"u}botter, Jonas and L{\"u}beck, Frederike and Behric, Lejs and Baumann, Anton and Bagatella, Marco and Marta, Daniel and Hakimi, Ido and Shenfeld, Idan and Buening, Thomas Kleine and Guestrin, Carlos and others},
  journal={arXiv preprint arXiv:2601.20802},
  year={2026}
}

@article{yang2026self,
  title={Self-distilled rlvr},
  author={Yang, Chenxu and Qin, Chuanyu and Si, Qingyi and Chen, Minghui and Gu, Naibin and Yao, Dingyu and Lin, Zheng and Wang, Weiping and Wang, Jiaqi and Duan, Nan},
  journal={arXiv preprint arXiv:2604.03128},
  year={2026}
}

@article{wang2026skill,
  title={Skill-sd: Skill-conditioned self-distillation for multi-turn llm agents},
  author={Wang, Hao and Wang, Guozhi and Xiao, Han and Zhou, Yufeng and Pan, Yue and Wang, Jichao and Xu, Ke and Wen, Yafei and Ruan, Xiaohu and Chen, Xiaoxin and others},
  journal={arXiv preprint arXiv:2604.10674},
  year={2026}
}

@article{yang2026opid,
  title={Opid: On-policy skill distillation for agentic reinforcement learning},
  author={Yang, Shuo and Wu, Jinyang and Lu, Zhengxi and Shen, Yuhao and Zhang, Fan and Feng, Lang and Zhang, Shuai and Luo, Haoran and Lian, Zheng and Wen, Zhengqi and others},
  journal={arXiv preprint arXiv:2606.26790},
  year={2026}
}

@article{wu2026seed,
  title={SEED: Self-Evolving On-Policy Distillation for Agentic Reinforcement Learning},
  author={Wu, Jinyang and Yang, Shuo and Lu, Zhengxi and Zhang, Fan and Shen, Yuhao and Feng, Lang and Luo, Haoran and Lian, Zheng and Zhang, Shuai and Wen, Zhengqi and others},
  journal={arXiv preprint arXiv:2607.14777},
  year={2026}
}

@article{zheng2026trajectory,
  title={Trajectory-Relative Hindsight Distillation for Agentic Reinforcement Learning},
  author={Zheng, Haoyu and Zhu, Yun and Wang, Qing and Zhang, Wenqiao},
  journal={arXiv preprint arXiv:2608.07371},
  year={2026}
}

@article{lan2026trust,
  title={Trust Is Not Enough: Influence Calibration for On-Policy Self-Distillation in Agentic RL},
  author={Lan, Qizhen and Xiao, Xi and Guan, Xiangchen and Fan, Mengchen and Lin, Moule and Choi, Jung Im and Zhu, Lijing},
  journal={arXiv preprint arXiv:2608.14945},
  year={2026}
}

@article{qu2026turnsight,
  title={TurnSight: Turn-Level Hindsight Self-Distillation for Tool-Integrated Reasoning},
  author={Qu, Changle and Dai, Sunhao and Cai, Hengyi and Zhou, Yuqi and Chen, Xinran and Xu, Jun and others},
  journal={arXiv preprint arXiv:2608.04007},
  year={2026}
}

@inproceedings{wang2024math,
  title={Math-shepherd: Verify and reinforce llms step-by-step without human annotations},
  author={Wang, Peiyi and Li, Lei and Shao, Zhihong and Xu, Runxin and Dai, Damai and Li, Yifei and Chen, Deli and Wu, Yu and Sui, Zhifang},
  booktitle={Proceedings of the 62nd Annual Meeting of the Association for Computational Linguistics (Volume 1: Long Papers)},
  pages={9426--9439},
  year={2024}
}

@article{luo2024improve,
  title={Improve mathematical reasoning in language models by automated process supervision},
  author={Luo, Liangchen and Liu, Yinxiao and Liu, Rosanne and Phatale, Samrat and Guo, Meiqi and Lara, Harsh and Li, Yunxuan and Shu, Lei and Zhu, Yun and Meng, Lei and others},
  journal={arXiv preprint arXiv:2406.06592},
  year={2024}
}

@article{cui2025process,
  title={Process reinforcement through implicit rewards},
  author={Cui, Ganqu and Yuan, Lifan and Wang, Zefan and Wang, Hanbin and Zhang, Yuchen and Chen, Jiacheng and Li, Wendi and He, Bingxiang and Fan, Yuchen and Yu, Tianyu and others},
  journal={arXiv preprint arXiv:2502.01456},
  year={2025}
}

@article{kazemnejad2024vineppo,
  title={Vineppo: Refining credit assignment in rl training of llms},
  author={Kazemnejad, Amirhossein and Aghajohari, Milad and Portelance, Eva and Sordoni, Alessandro and Reddy, Siva and Courville, Aaron and Roux, Nicolas Le},
  journal={arXiv preprint arXiv:2410.01679},
  year={2024}
}

@article{lu2026self,
  title={Self-distilled agentic reinforcement learning},
  author={Lu, Zhengxi and Yao, Zhiyuan and Han, Zhuowen and Wang, Zi-Han and Wu, Jinyang and Gu, Qi and Cai, Xunliang and Lu, Weiming and Xiao, Jun and Zhuang, Yueting and others},
  journal={arXiv preprint arXiv:2605.15155},
  year={2026}
}

@article{ye2026opdsearch+,
  title={OPDSearch+: On-Policy Distillation with RL Refinement for Search-Augmented Reasoning},
  author={Ye, Qinglin and Gu, Zhiyuan and Xia, Jingjie and Zhang, Yiheng and Zhao, Kaiyan and Zheng, Shunchao and Mu, Yuhang and Du, Wenchao and Wang, Yiming},
  journal={arXiv preprint arXiv:2608.24310},
  year={2026}
}

@article{li2026sequential,
  title={Sequential Beats Joint: On the Interplay between On-Policy Distillation and RLVR},
  author={Li, Boyan and Chen, Bingsen and Yang, Chenghao and Nie, Ping and Zhao, Chen and Ye, Xi},
  journal={arXiv preprint arXiv:2609.04108},
  year={2026}
}

@article{kim2026opsd,
  title={OPSD Compresses What RLVR Teaches: A Post-RL Compaction Stage for Reasoning Models},
  author={Kim, Jaehoon and Lee, Dongha},
  journal={arXiv preprint arXiv:2605.06188},
  year={2026}
}

@inproceedings{liu2023g,
  title={G-eval: NLG evaluation using gpt-4 with better human alignment},
  author={Liu, Yang and Iter, Dan and Xu, Yichong and Wang, Shuohang and Xu, Ruochen and Zhu, Chenguang},
  booktitle={Proceedings of the 2023 conference on empirical methods in natural language processing},
  pages={2511--2522},
  year={2023}
}

@article{dunn2017searchqa,
  title={Searchqa: A new q\&a dataset augmented with context from a search engine},
  author={Dunn, Matthew and Sagun, Levent and Higgins, Mike and Guney, V Ugur and Cirik, Volkan and Cho, Kyunghyun},
  journal={arXiv preprint arXiv:1704.05179},
  year={2017}
}

@article{lewis2020retrieval,
  title={Retrieval-augmented generation for knowledge-intensive nlp tasks},
  author={Lewis, Patrick and Perez, Ethan and Piktus, Aleksandra and Petroni, Fabio and Karpukhin, Vladimir and Goyal, Naman and K{\"u}ttler, Heinrich and Lewis, Mike and Yih, Wen-tau and Rockt{\"a}schel, Tim and others},
  journal={Advances in neural information processing systems},
  volume={33},
  pages={9459--9474},
  year={2020}
}

@article{wang2022text,
  title={Text embeddings by weakly-supervised contrastive pre-training},
  author={Wang, Liang and Yang, Nan and Huang, Xiaolong and Jiao, Binxing and Yang, Linjun and Jiang, Daxin and Majumder, Rangan and Wei, Furu},
  journal={arXiv preprint arXiv:2212.03533},
  year={2022}
}

@article{huang2026bicaa,
  title={BiCAA: Bidirectional Credit Assignment for Search-Augmented Agent},
  author={Huang, Yibin and Xu, Bin and Cao, Hailong and Zhu, Conghui},
  journal={arXiv preprint arXiv:2608.01321},
  year={2026}
}

\appendix
\section{Supplementary Implementation Details}
\label{sec:implementation-details}

Section~\ref{sec:setup} gives the shared training configuration. Here we specify additional implementation details. We use one PPO epoch per update, a dual-clip bound of $3$, and no reference-policy KL penalty.

\paragraph{Teacher construction and performance.}
For every benchmark, the teacher is a frozen Qwen2.5-7B-Instruct model trained with GRPO using the same task environment, prompts, action parser, reward function, rollout configuration, optimization hyperparameters, and training budget as the corresponding student run. The only difference is backbone scale: the teacher uses 7B parameters, whereas the student uses Qwen2.5-Instruct at 1.5B, 3B, or 7B parameters. The GRPO-trained teacher is frozen throughout \tide training, used only to compute on-policy token log-probabilities, and never used at evaluation. We use the final checkpoint after the fixed training budget, without validation- or test-set-based checkpoint selection. Its task performance is reported by the 7B GRPO row in Table~\ref{tab:controlled-main}. For 1.5B and 3B students, the teacher is strictly larger in parameter count. For the 7B student, teacher and student have the same backbone scale; in this case, the teacher is a frozen task-trained GRPO policy rather than a larger-model teacher.

\paragraph{OPD signal implementation.}
For each student-sampled token, we subtract the behavior-policy log-probability from the frozen teacher's log-probability. Both quantities are fixed when forming the PPO advantage, so gradients flow only through the current-policy importance ratio. The resulting OPD term is a sampled-action distillation signal.

\subsection{Baseline Descriptions}
\label{sec:baseline-descriptions}

\paragraph{Vanilla.}
Zero-shot prompting with the base Qwen2.5-Instruct model. The model receives only the task description and environment observations, with no demonstrations or structured action guidance.

\paragraph{GRPO.}
Group Relative Policy Optimization~\citep{shao2024deepseekmath} trains the student with outcome-based reward, normalizing advantages across the group of rollouts sampled for each prompt. No teacher supervision is used.

\paragraph{GiGPO.}
Group-in-Group Policy Optimization~\citep{feng2026group} supplements GRPO with a step-level relative advantage obtained by grouping matched interaction contexts within each prompt.

\paragraph{OPD.}
On-policy distillation~\citep{agarwal2024policy} uses the frozen GRPO-trained teacher's log-probability on student-sampled tokens as an additional PPO advantage signal. It is the distillation-only baseline in Table~\ref{tab:controlled-main} and uses the same frozen task-trained teacher and training configuration as ATOD and \tide at each student scale.

\paragraph{OPSD.}
On-Policy Self-Distillation~\citep{zhao2026self} reinforces the student's verified on-policy solutions through self-distillation. It uses no frozen external teacher and serves as a self-improving distillation reference.

\paragraph{ATOD.}
Annealed Turn-Aware On-Policy Distillation~\citep{tan2026atod} uses a prescribed linear OPD-to-RL annealing schedule. We reproduce ATOD under the same task environment, prompts, action parser, reward function, rollout configuration, optimization hyperparameters, training budget, decoding settings, and evaluation protocol as \tide. Its values in Table~\ref{tab:controlled-main} are mean $\pm$ sample standard deviation over three independently trained seeds.

\paragraph{SDAR.}
Self-Distilled Agentic Reinforcement Learning~\citep{lu2026self} combines distillation and RL through per-token gating within a shared student--teacher policy. We report its available 3B and 7B results as a method-specific hybrid reference.

\subsection{Benchmark-specific Evaluation Protocol}
Table~\ref{tab:controlled-main} uses ALFWorld's official valid-seen split (140 tasks); its Avg.\ is the task-count-weighted success rate across the six task types. The 134-task valid-unseen split is reported only for controlled ablations and sensitivity analyses, since corresponding records are unavailable for some external baselines. Other evaluation settings are given in Section~\ref{sec:setup}.

\subsection{GPT-5.5 Process-quality Diagnostic}
\label{sec:gpt-judge}

The GPT-5.5 scores in Table~\ref{tab:preexp2_gap_quality} are an external diagnostic only, following the use of LLM-based evaluators for fine-grained generation assessment~\citep{liu2023g}: they are not used in training, reward construction, checkpoint selection, hyperparameter selection, or test-time action selection. We randomly sample 100 full trajectories from the WebShop test set and use the trajectory, rather than the turn, as the sampling unit. We score every executable turn in the sampled trajectories.

\paragraph{Blind judge input.}
For each turn, the judge receives the task description, at most two preceding observation--action pairs, the current observation, the admissible actions, and the parsed executed action. It does \emph{not} receive the terminal reward, trajectory-success label, subsequent observations or actions, teacher/student log-probabilities, disagreement values, relative turn position, or the model's hidden reasoning. This restriction makes the diagnostic independent of the outcome quantities subsequently reported in Table~\ref{tab:preexp2_gap_quality}.

\paragraph{Judge prompt and score.}
We query GPT-5.5 once per turn. The judge evaluates the executed action rather than proposing an alternative and assigns an integer score on a $[1,3]$ scale. This coarse rubric is appropriate because the judge observes only the current decision context, rather than its future consequences. We do not use repeated judge calls or human annotations for this diagnostic, so it should be interpreted as supporting evidence for the motivation rather than a ground-truth process label. The following prompt is used verbatim, with bracketed fields instantiated for the evaluated turn:

\begin{center}
\setlength{\fboxsep}{7pt}
\colorbox{gray!12}{\begin{minipage}{0.94\linewidth}\small
\noindent\textbf{User.} Evaluate the \emph{executed action} of an autonomous agent in a WebShop shopping task using only the task, recent interaction history, current observation, admissible actions, and executed action below. Do not infer future observations, terminal reward, purchase outcome, trajectory success, teacher preference, student probability, disagreement statistic, or hidden reasoning. Judge the action itself, not its action type.

\noindent Apply the following criteria. For a search action, assess whether its query targets the task's explicit constraints. For a product click, assess whether the visible title, price, or metadata makes the product a reasonable candidate. For an option, description, feature, or review click, assess whether it selects or verifies a task-relevant attribute. For back, previous, or next actions, assess whether navigation is justified by visible evidence or needed to continue a reasonable search. For \texttt{Buy Now}, assign a positive score only if the visible information verifies all explicit task constraints, including the selected variant when applicable.

\noindent Assign one integer score: $1$ = clearly harmful, invalid, or inconsistent with the visible task constraints; $2$ = reasonable but inconclusive information gathering, verification, or navigation; $3$ = clearly advances the task based on visible evidence, or completes a purchase whose explicit constraints have been verified. When the visible context is insufficient to establish either an error or clear progress, assign $2$. Return only valid JSON: \texttt{\{``score'': <1, 2, or 3>, ``reason'': ``<one concise sentence>"\}}.

\noindent Task: [task description]. Recent interaction history: [up to two preceding observation--action pairs]. Current observation: [current observation]. Admissible actions: [admissible actions]. Executed action: [parsed executed action].
\end{minipage}}
\end{center}

\subsection{Relative Action Value Details}
\label{sec:local-details}

We follow GiGPO's relative process-value comparison~\citep{feng2026group} with exact recent-history matching. The resulting relative action value $Q_t$ is an outcome-conditioned credit heuristic over alternatives reached from matched recent states, related to rollout-based process-value estimation~\citep{cui2025process,kazemnejad2024vineppo}; it is not an independent causal attribution of a turn's contribution. Our configuration uses exact state-history matching with two preceding anchor states (history length $2$), mean-centered group scores, length-weight exponent $1$, and no trajectory-level base group.

\paragraph{Discounted return.}
Given a $T$-turn trajectory $\tau$ with task outcome $R(\tau)$ and a discount factor $\gamma$, the discounted return from turn $t$ onward is
\begin{equation}
G_t = \gamma^{T-t}\,R(\tau).
\end{equation}
We set $\gamma=0.95$ in all experiments. The discounted return is not used directly as a local weight: before it contributes to $Q_t$, it is mean-centered within a matched-history comparison group. This removes the group-level return baseline, but it does not algebraically remove the effect of discounting when matched continuations have different remaining lengths. Thus, $Q_t$ is not determined by raw turn position alone---it also depends on the alternatives sharing the same state history---but it is a time-discounted outcome-conditioned heuristic rather than position-free or causal turn attribution.

\paragraph{Exact recent observation-context grouping.}
Let $s_t$ be the environment-provided anchor observation at turn $t$, before prompt formatting or the addition of interaction history. The policy input retains the two most recent observation--action pairs, whereas process-value grouping uses exact suffixes of recent anchor observations. For prompt group $g$ and suffix length $k$, its comparison key is
\begin{equation}
K_{t,k}=\bigl(g,k,(s_{t-k+1},\ldots,s_t)\bigr),\qquad k\in\{1,2,3\}.
\end{equation}
Here $k=3$ corresponds to the current anchor state plus the two preceding states (the configured history length is $2$). We group turns with identical keys; textual states are matched by exact string equality, while arrays, lists, and dictionaries are recursively converted to hashable tuples. We do not use learned or semantic-similarity matching. Comparing downstream returns within a group yields a relative score from a matched recent context.

\paragraph{Per-group normalization.}
Within each group $K_{t,k}$ that contains at least two members, we compute a mean-centered process value:
\begin{equation}
p_{t,k} = G_t - \mu_{K_{t,k}}.
\end{equation}
This is the configured \texttt{mean\_norm} mode: we do not divide by the group standard deviation. $\mu_{K_{t,k}}$ is the mean discounted return in the group. Negative $p_{t,k}$ indicates a below-average outcome within the matched context, whereas positive $p_{t,k}$ indicates a better-than-average outcome; a group with no return variation yields only zero scores.

\paragraph{Aggregation across history lengths.}
A turn may have nonzero scores from multiple suffix lengths. We aggregate them using a length-weighted combination:
\begin{equation}
Q_t = \sum_{k \in \mathcal{H}_t} \frac{(k+1)^\nu}{\sum_{q \in \mathcal{H}_t}(q+1)^\nu}\,p_{t,k},
\end{equation}
where $\mathcal{H}_t\subseteq\{1,2,3\}$ contains suffix lengths with a nonzero centered score for turn $t$, and $\nu$ controls the preference for longer suffixes. We set $\nu=1$ and do not include a trajectory-level base group. A turn has no value-discriminative matched group when every exact-history group either contains only that turn or has no return variation; in this case, it is assigned raw $Q_t=0$.

The trajectory-wise normalization and the local score $z_t$ are defined in Equation~\ref{eq:local-score}. A turn with no value-discriminative matched group retains raw $Q_t=0$ and is excluded from the value min--max range; its value factor is set to one as a disagreement-only OPD fallback. Thus, it receives disagreement-only OPD modulation but no value-driven RL advantage. The same fallback applies when the valid matched turns have no within-set value range. Disagreement is normalized over all turns; when its range is at most $\epsilon$, its factor is one for every turn. All reported experiments use this fallback.

\paragraph{Local-ablation definitions.}
All local-allocation variants retain the global handoff, teacher, training budget, and combined OPD--RL objective. The full method uses relative action value as the RL advantage, normalized disagreement as the RL turn weight, and the product of normalized relative action value and disagreement as the OPD turn weight. The individual variants change these quantities as follows:
\begin{itemize}
\item \emph{w/o Process Reward}: removes relative action value from both branches. RL uses the unmodulated trajectory-level GRPO advantage; normalized disagreement is used as the turn weight for both RL and OPD.
\item \emph{w/o Disagreement}: removes normalized disagreement from both branches. RL retains relative action value as its advantage but uses a unit turn weight, whereas OPD uses normalized relative action value as its turn weight.
\item \emph{w/o OPD Mod.}: retains the full RL branch but replaces the OPD turn weight with a unit weight.
\item \emph{w/o RL Mod.}: retains the full OPD branch and relative-action-value RL advantage but replaces the RL turn weight with a unit weight.
\item \emph{Additive Fusion}: retains the RL branch and replaces the OPD turn weight with the arithmetic mean of normalized relative action value and normalized disagreement.
\end{itemize}
Equivalently, with $q_t=\qscore_t$, $d_t=\dscore_t$, $A^{\mathrm{RL}}_{t,j}=Q_t$, and $A^{\mathrm{OPD}}_{t,j}=\delta_{t,j}$ for the full method, the variants are
\begin{align*}
\textit{w/o Process Reward}:&\quad A^{\mathrm{RL}}_{t,j}=A^{\mathrm{GRPO}}_{t,j},\quad w_t^{\mathrm{RL}}=w_t^{\mathrm{OPD}}=d_t;\\
\textit{w/o Disagreement}:&\quad A^{\mathrm{RL}}_{t,j}=Q_t,\quad w_t^{\mathrm{RL}}=1,\quad w_t^{\mathrm{OPD}}=q_t;\\
\textit{w/o OPD Mod.}:&\quad w_t^{\mathrm{OPD}}=1;\qquad
\textit{w/o RL Mod.}:\quad w_t^{\mathrm{RL}}=1;\\
\textit{Additive Fusion}:&\quad w_t^{\mathrm{OPD}}=(q_t+d_t)/2,\quad w_t^{\mathrm{RL}}=d_t.
\end{align*}

\paragraph{Normalization pseudocode.}
\begin{algorithm}[H]
\caption{Trajectory-wise local-priority computation}
\label{alg:local-normalization}
\begin{algorithmic}[1]
\REQUIRE Turn values $Q_{1:T}$, valid-turn set $\mathcal{V}$, disagreements $D_{1:T}$, threshold $\epsilon$
\STATE $\widetilde{Q}_t\leftarrow1$ for $t\notin\mathcal{V}$
\IF{$\mathcal{V}=\emptyset$ or $\max_{t\in\mathcal{V}}Q_t-\min_{t\in\mathcal{V}}Q_t\leq\epsilon$}
    \STATE $\widetilde{Q}_t\leftarrow1$ for $t\in\mathcal{V}$
\ELSE
    \STATE $\widetilde{Q}_t\leftarrow(Q_t-\min_{u\in\mathcal{V}}Q_u)/(\max_{u\in\mathcal{V}}Q_u-\min_{u\in\mathcal{V}}Q_u+\epsilon)$ for $t\in\mathcal{V}$
\ENDIF
\STATE $\Delta_D\leftarrow\max_t D_t-\min_t D_t$
\IF{$\Delta_D\leq\epsilon$}
    \STATE $\widetilde{D}_t\leftarrow1$ for all $t$
\ELSE
    \STATE $\widetilde{D}_t\leftarrow(D_t-\min_u D_u)/(\Delta_D+\epsilon)$ for all $t$
\ENDIF
\STATE $z_t\leftarrow\widetilde{Q}_t\widetilde{D}_t$ for all $t$
\RETURN $z_{1:T}$
\end{algorithmic}
\end{algorithm}

\begin{figure}[H]
  \centering
  \includegraphics[width=0.88\linewidth]{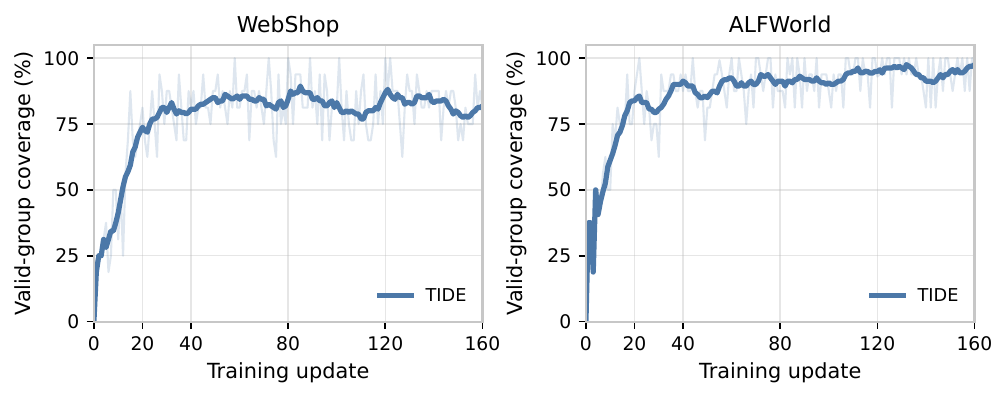}
  \caption{Valid matched-history coverage across 1.5B \tide training on WebShop and ALFWorld.}
  \label{fig:appendix-history-coverage}
\end{figure}

\paragraph{Coverage of matched-history comparisons.}
We measure valid-group coverage as the fraction of turns that have at least one exact recent observation-context group with two or more members and nonzero return variance. This measures how often relative action value is available for local allocation. Figure~\ref{fig:appendix-history-coverage} reports coverage for the 1.5B WebShop and ALFWorld \tide training runs. Coverage is lower during early exploration, when repeated contexts are sparse, and then quickly becomes high and stable on both benchmarks. Thus, unmatched turns are concentrated early rather than dominating local modulation throughout training. Faint lines show per-update coverage, and bold curves show 10-update moving averages.

\section{Supplementary Results and Analyses}
\label{sec:additional-analysis}

\subsection{Additional SearchQA Global-Handoff Evaluation}
\label{sec:searchqa-results}

We additionally evaluate the global handoff on SearchQA~\citep{dunn2017searchqa} with Qwen2.5-1.5B-Instruct students to assess whether the scheduling mechanism transfers to retrieval-augmented question answering~\citep{lewis2020retrieval,huang2026bicaa}. This experiment does not apply local modulation. SearchQA contains NQ, TriviaQA, PopQA, HotpotQA, 2WikiMultiHopQA, MuSiQue, and Bamboogle. Agents retrieve three passages from Wikipedia-2018 with E5-base-v2~\citep{wang2022text}, retain four history turns, and take at most five interaction steps. We report exact-match accuracy on normalized final \texttt{<answer>} spans. Each update uses 128 questions with 8 rollouts per question; PPO minibatches contain 512 trajectories. Results are mean $\pm$ sample standard deviation over three independently trained random seeds.

Table~\ref{tab:searchqa_results} compares GRPO, OPD, and \tide. \tide achieves the highest macro-average EM of 39.4\%, exceeding OPD by 3.1 points and outperforming it on six of seven subsets. The result indicates that the global handoff remains beneficial in this setting; it does not test the local module.

\suppressfloats[t]
\begin{table}[H]
\caption{SearchQA exact-match accuracy with Qwen2.5-1.5B-Instruct students (\%). Bold indicates the highest displayed student result.}
\label{tab:searchqa_results}
\centering
\scriptsize
\setlength{\tabcolsep}{3.6pt}
\renewcommand{\arraystretch}{1.05}
\resizebox{\textwidth}{!}{
\begin{tabular}{llcccccccc}
\toprule
\textbf{Method} & \textbf{Type} & \textbf{NQ} & \textbf{TriviaQA} & \textbf{PopQA} & \textbf{HotpotQA} & \textbf{2Wiki} & \textbf{MuSiQue} & \textbf{Bamboogle} & \textbf{Avg.} \\
\midrule
GRPO & RL & 14.8\tstd{1.1} & 28.3\tstd{1.6} & 21.1\tstd{0.7} & 15.8\tstd{1.4} & 23.9\tstd{1.2} & 2.7\tstd{0.1} & 9.3\tstd{0.8} & 16.6\tstd{0.3} \\
OPD & Distill & 36.0\tstd{0.8} & 49.1\tstd{1.4} & 38.8\tstd{0.6} & 36.8\tstd{1.1} & 36.2\tstd{0.9} & 26.1\tstd{1.7} & \textbf{31.2}\tstd{1.0} & 36.3\tstd{0.5} \\
\textbf{\tide} & Hybrid & \textbf{40.8}\tstd{1.2} & \textbf{51.5}\tstd{0.9} & \textbf{44.3}\tstd{0.5} & \textbf{42.4}\tstd{1.4} & \textbf{38.3}\tstd{1.0} & \textbf{27.7}\tstd{1.6} & 30.6\tstd{0.8} & \textbf{39.4}\tstd{0.3} \\
\bottomrule
\end{tabular}
}
\end{table}

\subsection{Global Handoff Analysis}
\label{sec:per-task-handoff}

\paragraph{Motivation and setup.}
The global mechanism uses discrepancy as a practical event signal rather than following a prescribed training-time schedule. To characterize the realized schedule, we trace the handoff state $r_k$ throughout training on WebShop, ALFWorld, and SearchQA.

\paragraph{Alternative schedule definitions.}
Success Handoff sets the handoff state $r_k$ in Equation~\ref{eq:global-coefficients} to the observed batch success rate. Fixed 1:1 Mixture keeps unit OPD and RL coefficients throughout training. Linear and Cosine Handoff use prescribed handoff states to interpolate the OPD and RL coefficients. With normalized training progress $s_k=(k-1)/(K-1)$ over $K$ updates, they set $r_k=s_k$ and $r_k=[1-\cos(\pi s_k)]/2$, respectively.

\begin{figure*}[t]
  \centering
  \includegraphics[width=0.88\textwidth]{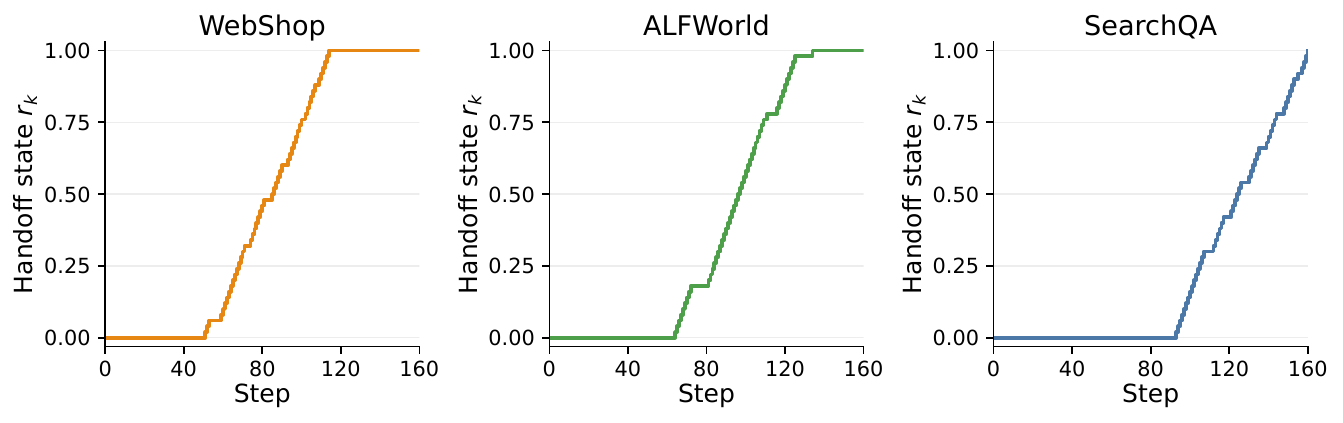}
  \caption{Global handoff states across WebShop, ALFWorld, and SearchQA.}
  \label{fig:appendix-coefficient-shift}
\end{figure*}

\paragraph{Results.}
Figure~\ref{fig:appendix-coefficient-shift} reports the realized task-level handoff trajectories. All begin with $r_k=0$ and exhibit task-dependent transition windows and intermediate plateaus. The plot documents how the chosen event rule adapts its transition timing across tasks.

\paragraph{Parameter sensitivity.}
\label{sec:handoff-sensitivity}

The plateau threshold $\zeta$ determines when a handoff step is triggered, and the step size $\eta$ determines its magnitude. To test whether the handoff depends on a narrow parameter choice, we vary each independently across a $5\times$ range using Qwen2.5-1.5B-Instruct students on WebShop and ALFWorld. All other settings follow the main experimental configuration.

\begingroup
\renewcommand{\tstd}[1]{\,$\scriptstyle\pm #1$}
\begin{table}[H]
\caption{Sensitivity of global handoff parameters (\%).}
\label{tab:handoff-sensitivity}
\centering
\begin{minipage}[t]{0.485\textwidth}
\centering
(a) Plateau threshold $\zeta$\\[4pt]
\scriptsize
\setlength{\tabcolsep}{2.5pt}
\renewcommand{\arraystretch}{1.10}
\begin{tabular*}{\linewidth}{@{\extracolsep{\fill}}>{\centering\arraybackslash}m{0.20\linewidth}cccc@{}}
\toprule
\multirow{2}{*}{\raisebox{-0.5ex}{$\zeta$}} & \multicolumn{2}{c}{\textbf{WebShop}} & \multicolumn{2}{c}{\textbf{ALFWorld}} \\
\cmidrule(lr){2-3}\cmidrule(lr){4-5}
& \mbox{SR $\uparrow$} & \mbox{Score $\uparrow$} & \mbox{Seen $\uparrow$} & \mbox{Unseen $\uparrow$} \\
\midrule
0.01 & 76.0\tstd{1.9} & 89.1\tstd{1.2} & 85.7\tstd{1.5} & 84.3\tstd{1.7} \\
0.02 & \textbf{77.2}\tstd{1.8} & \textbf{89.8}\tstd{1.2} & \textbf{86.0}\tstd{0.8} & \textbf{85.1}\tstd{1.5} \\
0.05 & 75.6\tstd{2.0} & 88.4\tstd{1.3} & 85.0\tstd{1.6} & 82.8\tstd{1.9} \\
\bottomrule
\end{tabular*}
\end{minipage}
\hfill
\begin{minipage}[t]{0.485\textwidth}
\centering
(b) Handoff step size $\eta$\\[4pt]
\scriptsize
\setlength{\tabcolsep}{2.5pt}
\renewcommand{\arraystretch}{1.10}
\begin{tabular*}{\linewidth}{@{\extracolsep{\fill}}>{\centering\arraybackslash}m{0.20\linewidth}cccc@{}}
\toprule
\multirow{2}{*}{\raisebox{-0.5ex}{$\eta$}} & \multicolumn{2}{c}{\textbf{WebShop}} & \multicolumn{2}{c}{\textbf{ALFWorld}} \\
\cmidrule(lr){2-3}\cmidrule(lr){4-5}
& \mbox{SR $\uparrow$} & \mbox{Score $\uparrow$} & \mbox{Seen $\uparrow$} & \mbox{Unseen $\uparrow$} \\
\midrule
0.01 & 76.4\tstd{1.9} & 89.2\tstd{1.2} & 85.7\tstd{1.5} & 83.6\tstd{1.8} \\
0.02 & \textbf{77.2}\tstd{1.8} & \textbf{89.8}\tstd{1.2} & \textbf{86.0}\tstd{0.8} & \textbf{85.1}\tstd{1.5} \\
0.05 & 75.8\tstd{2.0} & 88.6\tstd{1.3} & 85.0\tstd{1.7} & 82.1\tstd{2.0} \\
\bottomrule
\end{tabular*}
\end{minipage}
\end{table}
\endgroup

\paragraph{Results.}
The default setting, $\zeta=\eta=0.02$, gives the strongest result in each reported column. Performance remains stable over the evaluated range: all variants remain within 1.6 points on WebShop SR and 3.0 points on ALFWorld unseen, indicating robustness to moderate changes in either parameter.

\subsection{Additional 3B Ablations}
\label{sec:ablation-3b}

\paragraph{Motivation and setup.}
We repeat the global-handoff and local-allocation ablations with Qwen2.5-3B-Instruct students to test whether the 1.5B findings persist at a larger student scale. Each variant uses the same teacher, rollout budget, training budget, and evaluation protocol as the 3B main result; global variants differ only in their handoff schedule, and local variants differ only in their turn-level allocation.

\begingroup
\renewcommand{\tstd}[1]{\,$\scriptstyle\pm #1$}
\begin{table}[t]
\caption{TIDE ablations with Qwen2.5-3B-Instruct students (\%). Bold indicates the highest displayed value.}
\label{tab:ablation-3b}
\centering
\begin{minipage}[t]{0.485\textwidth}\centering
(a) Global handoff strategies\\[4pt]
\scriptsize\setlength{\tabcolsep}{1.8pt}\renewcommand{\arraystretch}{1.10}
\begin{tabular*}{\linewidth}{@{\extracolsep{\fill}}>{\centering\arraybackslash}m{0.30\linewidth}cccc@{}}
\toprule
\multirow{2}{*}{\raisebox{-0.5ex}{\textbf{Method}}} & \multicolumn{2}{c}{\textbf{WebShop}} & \multicolumn{2}{c}{\textbf{ALFWorld}} \\
\cmidrule(lr){2-3}\cmidrule(lr){4-5}
& \mbox{\textbf{SR $\uparrow$}} & \mbox{\textbf{Score $\uparrow$}} & \mbox{\textbf{Seen $\uparrow$}} & \mbox{\textbf{Unseen $\uparrow$}} \\
\midrule
GRPO & 63.3\tstd{2.0} & 79.8\tstd{1.1} & 74.0\tstd{0.8} & 60.4\tstd{2.2} \\
Success Handoff & 74.2\tstd{2.3} & 84.5\tstd{1.4} & 79.3\tstd{1.9} & 77.6\tstd{2.2} \\
Fixed 1:1 Mixture & 74.9\tstd{1.8} & 85.8\tstd{1.0} & 82.1\tstd{1.2} & 80.3\tstd{1.7} \\
Linear Handoff & 75.2\tstd{2.1} & 87.5\tstd{1.2} & 87.1\tstd{1.4} & 84.6\tstd{1.7} \\
Cosine Handoff & 76.0\tstd{1.6} & 88.4\tstd{0.9} & 86.4\tstd{1.4} & 85.1\tstd{1.5} \\
\tide & \textbf{79.0}\tstd{1.5} & \textbf{90.2}\tstd{0.9} & \textbf{89.3}\tstd{0.7} & \textbf{87.3}\tstd{1.5} \\
\bottomrule\end{tabular*}\end{minipage}\hfill
\begin{minipage}[t]{0.485\textwidth}\centering
(b) Local signal allocation\\[4pt]
\scriptsize\setlength{\tabcolsep}{1.8pt}\renewcommand{\arraystretch}{1.10}
\begin{tabular*}{\linewidth}{@{\extracolsep{\fill}}>{\centering\arraybackslash}m{0.30\linewidth}cccc@{}}
\toprule
\multirow{2}{*}{\raisebox{-0.5ex}{\textbf{Method}}} & \multicolumn{2}{c}{\textbf{WebShop}} & \multicolumn{2}{c}{\textbf{ALFWorld}} \\
\cmidrule(lr){2-3}\cmidrule(lr){4-5}
& \mbox{\textbf{SR $\uparrow$}} & \mbox{\textbf{Score $\uparrow$}} & \mbox{\textbf{Seen $\uparrow$}} & \mbox{\textbf{Unseen $\uparrow$}} \\
\midrule
w/o OPD Mod. & 77.1\tstd{1.4} & 88.0\tstd{0.8} & 87.1\tstd{1.2} & 85.6\tstd{1.6} \\
w/o RL Mod. & 75.2\tstd{2.1} & 86.8\tstd{1.3} & 87.9\tstd{1.4} & 83.8\tstd{1.9} \\
w/o Process Reward & 74.8\tstd{2.3} & 86.2\tstd{1.5} & 83.6\tstd{1.9} & 83.1\tstd{2.2} \\
w/o Disagreement & 77.0\tstd{1.5} & 87.4\tstd{1.0} & 87.1\tstd{1.4} & 84.6\tstd{1.6} \\
Additive Fusion & 70.3\tstd{3.4} & 77.5\tstd{2.5} & 74.3\tstd{2.1} & 73.1\tstd{2.2} \\
\tide & \textbf{79.0}\tstd{1.5} & \textbf{90.2}\tstd{0.9} & \textbf{89.3}\tstd{0.7} & \textbf{87.3}\tstd{1.5} \\
\bottomrule\end{tabular*}\end{minipage}
\end{table}
\endgroup

\paragraph{Results.}
For global handoff, \tide reaches 79.0\% WebShop SR and 87.3\% ALFWorld unseen success, exceeding the strongest preset schedule, Cosine Handoff, by 3.0 and 2.2 points, respectively. For local allocation, it again outperforms every variant. Removing process reward, disagreement, OPD modulation, or RL modulation lowers WebShop SR to 74.8\%, 77.0\%, 77.1\%, and 75.2\%, respectively; Additive Fusion is lowest at 70.3\%. These results reproduce the two 1.5B findings at 3B: the discrepancy-triggered handoff improves over the evaluated schedules, and the joint local allocation is stronger than its component variants.

\end{document}